\documentclass[letterpaper, 10 pt, journal, twoside]{IEEEtran}
\ifCLASSINFOpdf
\else
\fi

\usepackage{graphics} 
\usepackage{epsfig} 
\usepackage{amsmath} 
\usepackage{amssymb}  
\usepackage{ifpdf}
\usepackage{algorithm}
\usepackage{algpseudocode}
\usepackage{multirow}
\usepackage[hidelinks]{hyperref}
\usepackage{fancyhdr}
\usepackage[caption=false, font=footnotesize]{subfig}
\usepackage{xcolor}

\newcommand{\blacktext}[1]{\textcolor{black}{#1}}

\usepackage{soul}

\usepackage{booktabs}       
\usepackage{threeparttable} 
\usepackage{balance}

\begin{document}
\title{Seam-to-Graph Reconstruction for Garment Configuration Alignment}

\author{Xuzhao~Huang$^{1}$,~\IEEEmembership{Graduate~Student~Member,~IEEE}, \\Kai~Tang$^{1}$,~\IEEEmembership{Graduate~Student~Member,~IEEE}, Fuyuki~Tokuda$^{2}$,~\IEEEmembership{Member,~IEEE}, Norman~C.~Tien$^{3}$,~\IEEEmembership{Senior~Member,~IEEE},~and~Kazuhiro~Kosuge$^{1}$,~\IEEEmembership{Life~Fellow,~IEEE}
    \thanks{$^{1}$Xuzhao Huang, Kai Tang, and Kazuhiro Kosuge are with the JC STEM Lab of Robotics for Soft Materials, Department of Electrical and Electronic Engineering, Faculty of Engineering, The University of Hong Kong, Hong Kong SAR, 000000, China (e-mail: x.z.huang@connect.hku.hk).}
    \thanks{$^{2}$Fuyuki Tokuda is with the Unprecedented-scale Data Analytics Center, Tohoku University, Sendai 980-0845, Japan, and also with the Graduate School of Information Sciences, Tohoku University, Sendai 980-0845, Japan.}
    \thanks{$^{3}$Norman C. Tien is with the Department of Electrical and Electronic Engineering, Faculty of Engineering, The University of Hong Kong, Hong Kong SAR, 000000, China. }%
}

\maketitle

\thispagestyle{fancy}
\fancyhf{}
\fancyhead[L]{\tiny\textbf{THIS WORK HAS BEEN SUBMITTED TO THE IEEE FOR POSSIBLE PUBLICATION. COPYRIGHT MAY BE TRANSFERRED WITHOUT NOTICE, AFTER WHICH THIS VERSION MAY NO LONGER BE ACCESSIBLE.}}

\begin{abstract}
    Seams encode rich structural information about garments but are frequently partially observable in robotic manipulation scenarios. To robustly leverage seam information, we propose a Seam-to-Graph network based on graph neural networks and attention mechanisms. This network maps unstructured seam observations to a topology-encoded structural skeleton graph for real-time garment state estimation.
    Using this skeleton-graph-based state estimation, we design a deformation-aware, hierarchical visual servoing controller for garment configuration alignment.
    We implement this controller on a bimanual robot system to load a garment onto a screen printing platen and to align it to the desired configuration precisely.
    Real-robot experiments demonstrate that the robot using the proposed method not only achieves human-level alignment accuracy with reduced variance in alignment error but is also robust to different garments.
    These results demonstrate that the use of seam information is effective for garment manipulation.
\end{abstract}


\IEEEpeerreviewmaketitle

\section{Introduction}\label{sec:introduction}
\begin{figure}[ht]
    \includegraphics[width=\linewidth]{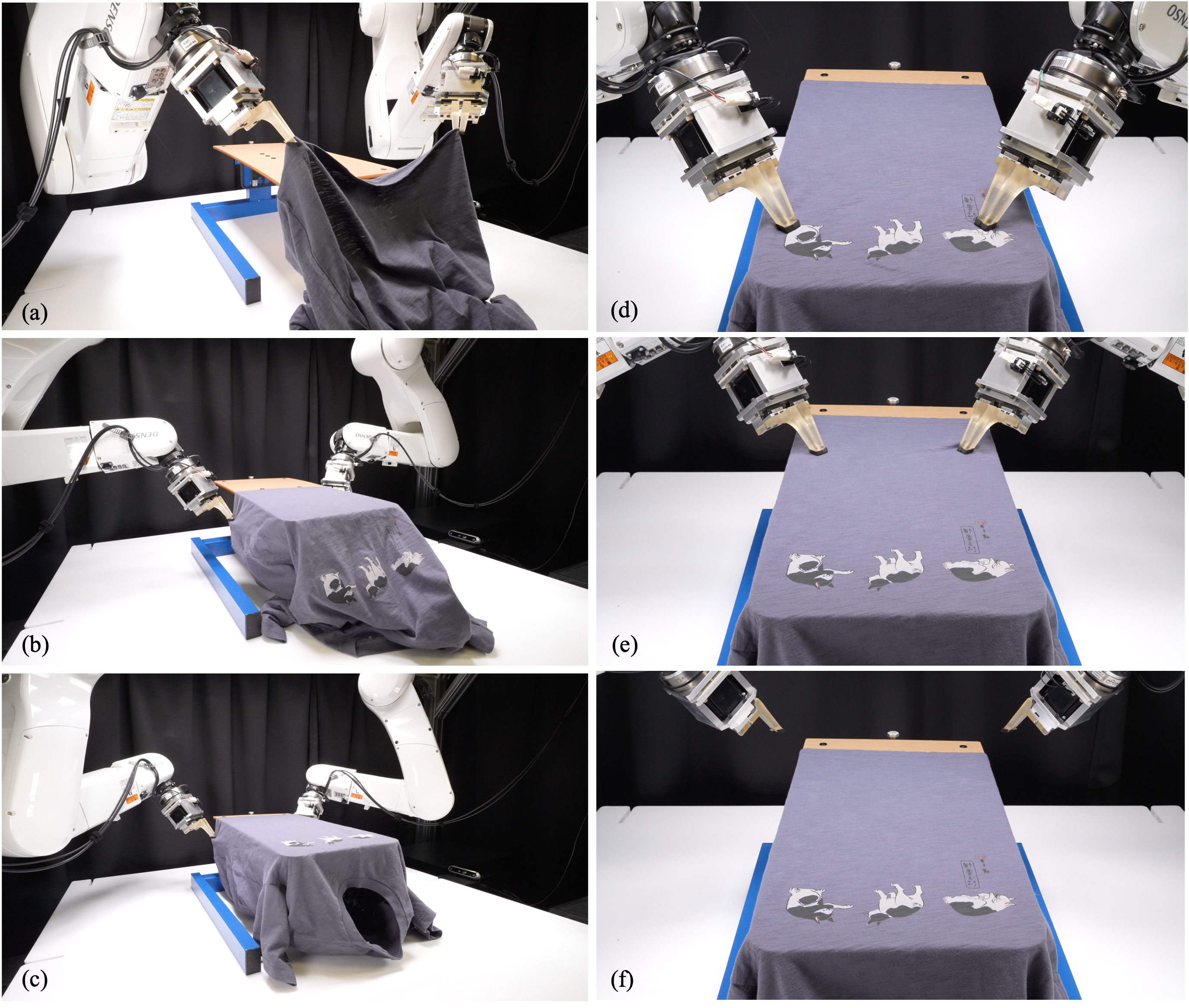}
    \caption{
        Example processes of garment configuration alignment, including (a) to (c) garment loading onto a platen, (d) and (e) fine alignment towards the desired configuration, and (f) the aligned result.
    }
    \label{fig:cover}
\end{figure}

Garment manipulation remains challenging in robotics due to its high degree of freedom, complex deformation, and frequent self-occlusion of garments.
Existing methods have demonstrated impressive progress in garment manipulation tasks such as folding~\cite{maitin2010towerfolding, wxm22ral, liu2025topological}, unfolding~\cite{ch26tase, sis}, dressing~\cite{dressing}, ironing~\cite{zyk25tase}, and hanging~\cite{graphgarment, graspcollar}. 

Most existing studies focus on achieving task-level success by roughly aligning the configuration of garments. However, garment manufacturing processes such as screen printing and sewing require robots to align garments or fabrics precisely and repeatedly to a prescribed configuration. This differs from the task-level success based on rough alignment because even small residual misalignment can affect the quality of the garment.

Precise manipulation of garments requires minimizing the discrepancy between their current and desired configurations. Minimizing these configurations requires a state representation that allows for comparison of the desired and current garment configurations.

Garment state representations can be categorized as follows~\cite{ai2025review}: 2D pixel representations~\cite{ch26tase, graspcollar}, latent representations~\cite{pi0}, 3D particle representations (point clouds~\cite{dressing}, meshes~\cite{garmentnets, edmund25icma, tk2025arxiv}), and keypoint representations~\cite{skeletonpoints}. 
These representations trade off estimation difficulty against reliability. Less-structured representations, such as pixels and latent states, simplify state estimation, but they introduce model bias and degrade downstream control. Keypoints are more compact yet sensitive to occlusion. Particle-based methods require low-noise 3D points, which are difficult to obtain. 

For precise garment configuration alignment tasks, a control-oriented state representation with task-specific simplification that explicitly encodes structural information of garments is necessary.

Seams are natural structural cues for estimating the state of garments. They encode the structural information of the garment since garments are constructed from multiple distinct panels of fabric stitched together.
However, raw seam observations cannot be used directly to determine the state of a garment because they are often partial, unordered, and noisy.

To address these problems, in this paper, we propose a Seam-to-Graph reconstruction method for garment state estimation to align the garment configuration precisely with the desired configuration. This method maps partial and unordered seam observations, together with point cloud observations, into a structural skeleton graph that is specific to each type of garment. The structural skeleton graph provides fixed graph topology for garment and compact geometric information for real-time state estimation and feedback control. For clarity of presentation, we refer to this structural skeleton graph as either \textbf{structural skeleton} or \textbf{skeleton} in this paper. 

Based on the reconstructed skeleton, we align the garment configuration to the desired one by aligning the coordinate of skeleton vertices. Since garments cannot be treated as single rigid objects, we develop a deformation-aware, hierarchical visual servoing controller that corrects local alignment errors iteratively.
This closed-loop mechanism enables the precise configuration alignment of the garment using a bimanual robot system.
We validate our proposed method through real robot experiments, using a screen printing application as an example, as shown in Fig.~\ref{fig:cover}.

In summary, the main contributions of this paper are as follows.
\begin{itemize}
    \item We introduce a seam-based structural skeleton graph representation for control-oriented garment state estimation.
    \item We propose a Seam-to-Graph reconstruction network that maps partial and unordered seam observations into a complete structural skeleton graph.
    This enables real-time garment state estimation.
    \item We develop a deformation-aware, hierarchical visual servoing controller for precise garment configuration alignment.
    \item We demonstrate the garment loading and alignment for screen printing using a real bimanual robot system. The experiments show that the alignment accuracy of the proposed method achieves human-level accuracy with improved repeatability.
\end{itemize}

The remainder of this paper is organized as follows. 
Section II reviews related work. 
Section III states the problem addressed in this paper. 
Section IV details the proposed method. 
Section V presents the experimental setup and results analysis. 
Section VI concludes the paper.

\section{Related Work}\label{sec:relatedwork}

\subsection{Robotic Garment Configuration Alignment}

Researchers have conducted extensive research on garment configuration alignment for various applications.
In the simulator, dragging motions are planned for dragging the garment onto board-shaped tools, which are then applied to real-world ironing and printing tasks~\cite{zyk25tase}. 
The grasping problem prior to loading is solved by adjusting the grasping points with visual feedback to achieve a better initial grasp before aligning the garment configuration~\cite{jose22iros}. 
The flattening and aligning of a piece of fabric on a plane has been achieved using estimated mesh representations and control pipelines~\cite{edmund25icma, tk2025arxiv}. 

Unlike these research, our work addresses the challenge of precisely aligning the configuration of a garment to its desired 3D spatial state.

\subsection{Garment State Representation and Estimation}

A core challenge in garment handling is the representation of the garment state.
Garmentnets~\cite{garmentnets} maps point cloud observations to a canonical space using a classification method and projects them back onto a mesh representation.
Learning-based methods inherently use image pixels or point clouds as garment state representations to generate robot actions~\cite{ch26tase, dressing}.
Other methods define discrete or coarse states based on keypoints~\cite{skeletonpoints}, folds, or edges. 

Mesh and graph representations are also widely used.
Cloth-Splatting~\cite{clothsplatting} employs a Graph Neural Network~(GNN) to predict cloth mesh states and uses Gaussian Splatting~\cite{gaussiansplatting} for closed-loop updating of garment states under RGB image supervision.
GraphNeuralCloth~\cite{graphneuralcloth} uses a GNN to predict cloth deformation from historical mesh sequences.
GraphGarment~\cite{graphgarment} converts a point cloud into a graph using K-nearest neighbors and applies a GNN to predict the future point cloud state. 
In~\cite{liu2025topological}, the authors represent garment states via topological graphs obtained using semantic regions.

Despite this progress, robust and precise state estimation remains challenging, which is essential for high-precision tasks, such as garment configuration alignment.
Seams address this challenge by providing structural information to reconstruct a complete topology-encoded structural skeleton graph from partial observations. The observed seams constrain graph node positions, thereby improving state estimation accuracy.

\begin{figure*}[htbp]
    \includegraphics[width=\textwidth]{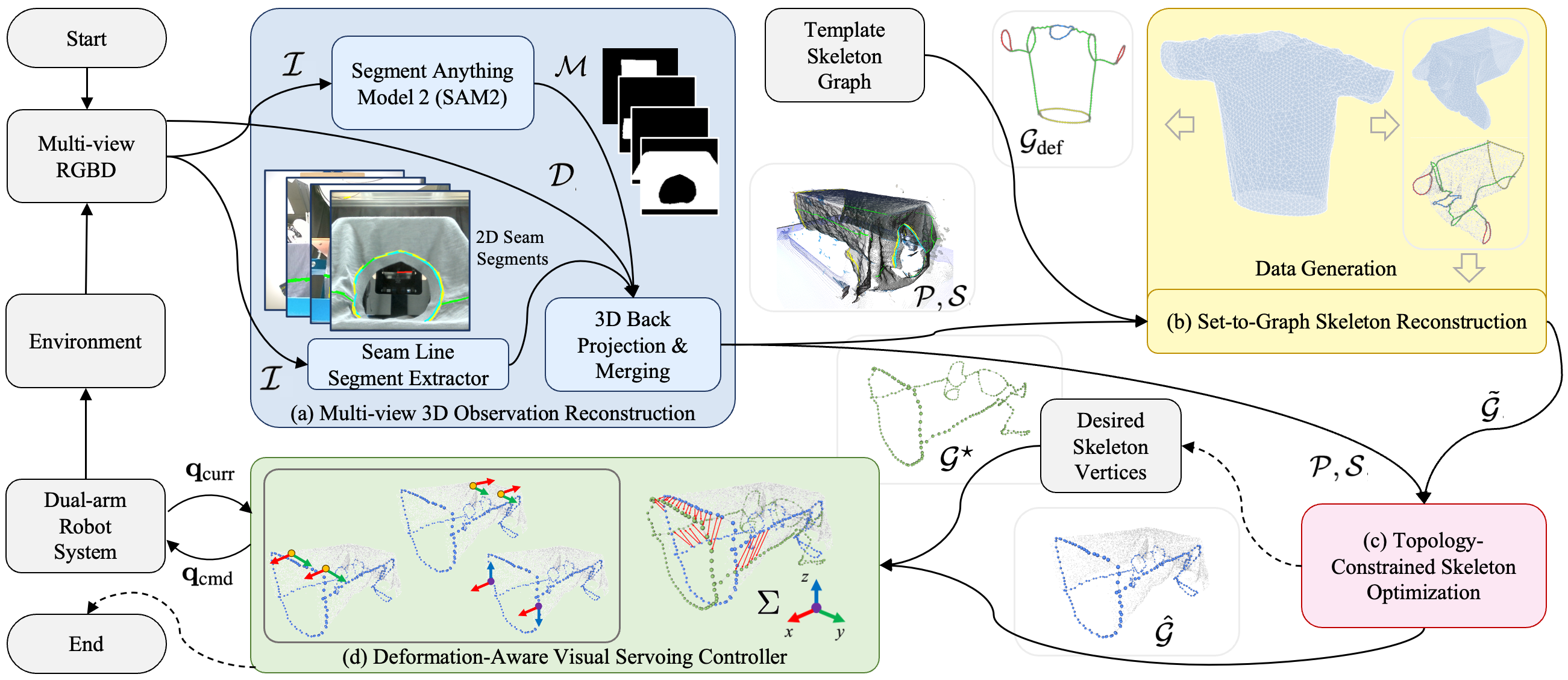}
    \caption{
        Overall pipeline. 
        Four modules are included to achieve garment loading and alignment using a bimanual robot.
        (a)~We use SAM2~\cite{sam2} and a seam line segment extractor~\cite{sis} to obtain garment masks $\mathcal{M}$ and 2D seam segments from multi-view RGB images $\mathcal{I}$.
        Back-projection yields a 3D point cloud $\mathcal{P}$ and 3D seam segments $\mathcal{S}$ using depth maps $\mathcal{D}$.
        (b)~A Seam-to-Graph network, trained on simulated data, predicts the structural skeleton $\tilde{\mathcal{G}}$ from $\mathcal{P}$, $\mathcal{S}$, and a template skeleton $\mathcal{G}_{\text{temp}}$.
        (c)~Topology-constrained skeleton optimization refines $\tilde{\mathcal{G}}$ using the observations $\mathcal{P}$, $\mathcal{S}$ to produce estimated skeleton $\hat{\mathcal{G}}$.
        (d)~A deformation-aware visual servoing controller computes the robot commands $\mathbf{q}_\text{cmd}$ w.r.t the world frame $\Sigma$ from $\hat{\mathcal{G}}$ and the goal skeleton $\mathcal{G}^{\star}$.
    }
    \label{fig:pipeline}
\end{figure*}

\subsection{Deformable Objects Manipulation}

Manipulating deformable objects is challenging due to the coupling between non-rigid motion and deformation.
Traditional visual servoing methods assume fixed geometric relationships between features, so they are designed for rigid objects~\cite{vs02tutorial, vs06ram}, which limits their direct application to deformable objects.

Reinforcement learning~\cite{Reinforcement} and imitation learning~\cite{tk2025arxiv} are used to train robot action policies for manipulating garments.
A neural network is proposed to select the Pick\&Place motion for flattening cloth~\cite{ch26tase}.

Beyond data-driven methods, model-based approaches have also been investigated.
Potential fields are combined with a parametric-curve-based shape servo for cable obstacle avoidance~\cite{Jiang2025}.

In this paper, we propose a deformation-aware visual servoing controller driven by the structural skeleton. 
Using a fixed-topology graph reconstructed from seam observations, our hierarchical controller aligns garment regions through quasi-rigid registration and position-based visual servoing. 
This enables robust and precise configuration alignment in real time.

\section{Problem Statement}\label{problemstatement}

We consider the task of autonomously loading a garment onto a rigid platen and precisely aligning it in the desired configuration using a bimanual robot system.
This task assumes that the two robotic end-effectors have properly grasped the garment initially.

To establish a garment state representation, we define a structural skeleton $\mathcal{G} = (\mathcal{V}, \mathcal{E}, \mathbf{X})$ as a geometric graph. $\mathcal{V}$ is a set of $N_{\text{vtx}}$ key vertices that lie on the garment's seams. The edges, $\mathcal{E} \subseteq \mathcal{V} \times \mathcal{V}$, encodes their fixed topological connectivity, and $\mathbf{X} \in \mathbb{R}^{N_{\text{vtx}} \times 3}$ is a matrix of their current 3D coordinates with respect to the world frame $\Sigma$.
In this formulation, $(\mathcal{V}, \mathcal{E})$ serves as a structural prior, and $\mathbf{X}$ provides the instantaneous geometric configuration.
Thus, the garment state is represented by $\mathcal{G}$.

Given preprocessed 3D structural observations of the garment in $\Sigma$, consisting of a point cloud $\mathcal{P}$ and 3D seam segments $\mathcal{S}$ extracted from raw RGB-D images, the perception subproblem is to estimate the real-time garment state $\hat{\mathcal{G}}$ by deforming a given template structural skeleton $\mathcal{G}_{\text{temp}}$ using a mapping $\mathcal{F}(\cdot)$:
\begin{equation}
    \hat{\mathcal{G}} = \mathcal{F}\bigl(\mathcal{P}, \mathcal{S}, \mathcal{G}_{\text{temp}}\bigr),
    \label{equ:perception}
\end{equation}
where $\hat{\mathcal{G}} = (\mathcal{V}, \mathcal{E}, \hat{\mathbf{X}})$ and $\mathcal{G}_{\text{temp}} = (\mathcal{V}, \mathcal{E}, \mathbf{X}_{\text{temp}})$ contain the estimated vertex coordinates $\hat{\mathbf{X}} \in \mathbb{R}^{N_{\text{vtx}} \times 3}$ and the template vertex coordinates $\mathbf{X}_{\text{temp}} \in \mathbb{R}^{N_{\text{vtx}} \times 3}$, respectively.
This mapping must robustly handle unordered and partially observed seams while strictly adhering to the topological constraints defined by $\mathcal{E}$.

Let the desired garment state be denoted by $\mathcal{G}^{\star} = (\mathcal{V}, \mathcal{E}, \mathbf{X}^{\star})$, where $\mathbf{X}^{\star} \in \mathbb{R}^{N_{\text{vtx}} \times 3}$ are the aligned vertex coordinates.
The control subproblem is to design a policy $\pi(\cdot)$ that drives the bimanual system to minimize the error between the current skeleton $\hat{\mathcal{G}}$ and the desired skeleton $\mathcal{G}^{\star}$.
The control law is formulated as:
\begin{equation}
    \mathbf{q}_{\text{cmd}} = \pi\bigl(\hat{\mathcal{G}}, \mathcal{G}^{\star}\bigr),
    \label{equ:control}
\end{equation}
where $\mathbf{q}_{\text{cmd}}$ represents the velocity or position commands sent to the robot.
A critical constraint in this subproblem is mitigating deformation coupling. The policy must guide each local region toward its desired condition to ensure the entire garment is aligned within the prescribed tolerances.

\section{Methodology}\label{sec:method}

This section details the proposed method for estimating the structural skeleton state and the corresponding control pipeline.
The overall workflow is shown in Fig.~\ref{fig:pipeline}.
The workflow consists of four sequential modules that reconstruct the structural skeleton graph $\hat{\mathcal{G}}$ from raw inputs and derive the control commands.

\subsection{Multi-view 3D Observation Reconstruction}

This section details the process of converting multi-view RGB-D images into the structured inputs, $\mathcal{P}$ and $\mathcal{S}$, required by Eq. (1), as shown in Fig. 1(a). This process connects raw sensor data to the formalized problem presented in Section~\ref{problemstatement}.

\subsubsection{3D Point Cloud Reconstruction}
The raw data from each camera consists of an RGB image $\mathcal{I}_k$ and a depth map $\mathcal{D}_k$.
The intrinsic matrix $\mathbf{K}_k$ and extrinsic matrix $[\mathbf{R}_k | \mathbf{t}_k]$ of camera $k$ are pre-calibrated.
For each view, the 3D coordinate of pixel $(u,v)$ expressed in the world frame $\Sigma$ is calculated as follows:
\begin{equation}
    \mathbf{p}_k(u,v) = \mathbf{R}_k \left( \mathbf{K}_k^{-1} [u, v, 1]^{\mathsf{T}} \cdot \mathcal{D}_k(u,v) \right) + \mathbf{t}_k.
    \label{equ:world_point}
\end{equation}

To remove background interference, we apply the Segment Anything Model~2 (SAM2)~\cite{sam2} for garment instance segmentation. This yields a pixel-level binary mask $\mathcal{M}_k$, where $\mathcal{M}_k(u,v)=1$ indicates that the pixel belongs to the garment.
By applying Eq.~\ref{equ:world_point} to all masked pixels and merging the results across views, we obtain a clean 3D point cloud of the garment:
\begin{equation}
    \mathcal{P} = \bigcup_{k=1}^{K} \left\{ \mathbf{p}_k(u_k,v_k) \;\middle|\; \mathcal{M}_k(u_k,v_k)=1 \right\}.
    \label{equ:pointcloudfusing}
\end{equation}

\subsubsection{3D Seam Segment Reconstruction}
We utilize a convolutional neural network-based method~\cite{sis, yolo11} to extract seam line segments from the RGB image $\mathcal{I}_k$.
Similar to~\cite{sis}, each extracted segment is assigned a visual category $c_{k,j}$ (such as solid line, dashed line, etc.), which corresponds to different sewing techniques and conveys structural semantics related to garment assembly.
Let $J_k$ denote the number of detected segments in view $k$.
Each segment $j$ is represented by the 2D image coordinates of its endpoints: $(u_{k,j}^1, v_{k,j}^1)$ and $(u_{k,j}^2, v_{k,j}^2)$, together with its visual category $c_{k,j}$.
Using Eq.~\ref{equ:world_point}, each 2D seam segment is back-projected into 3D space.
The 3D seam segments are merged across views into a set:
\begin{equation}
    \mathcal{S} = \bigcup_{k=1}^{K} \bigcup_{j=1}^{J_k} \left\{ \big( \mathbf{p}_k(u_{k,j}^1,v_{k,j}^1),\; \mathbf{p}_k(u_{k,j}^2,v_{k,j}^2),\; c_{k,j} \big) \right\}.
    \label{equ:seam_fusing}
\end{equation}

\begin{figure}
    \includegraphics[width=\linewidth]{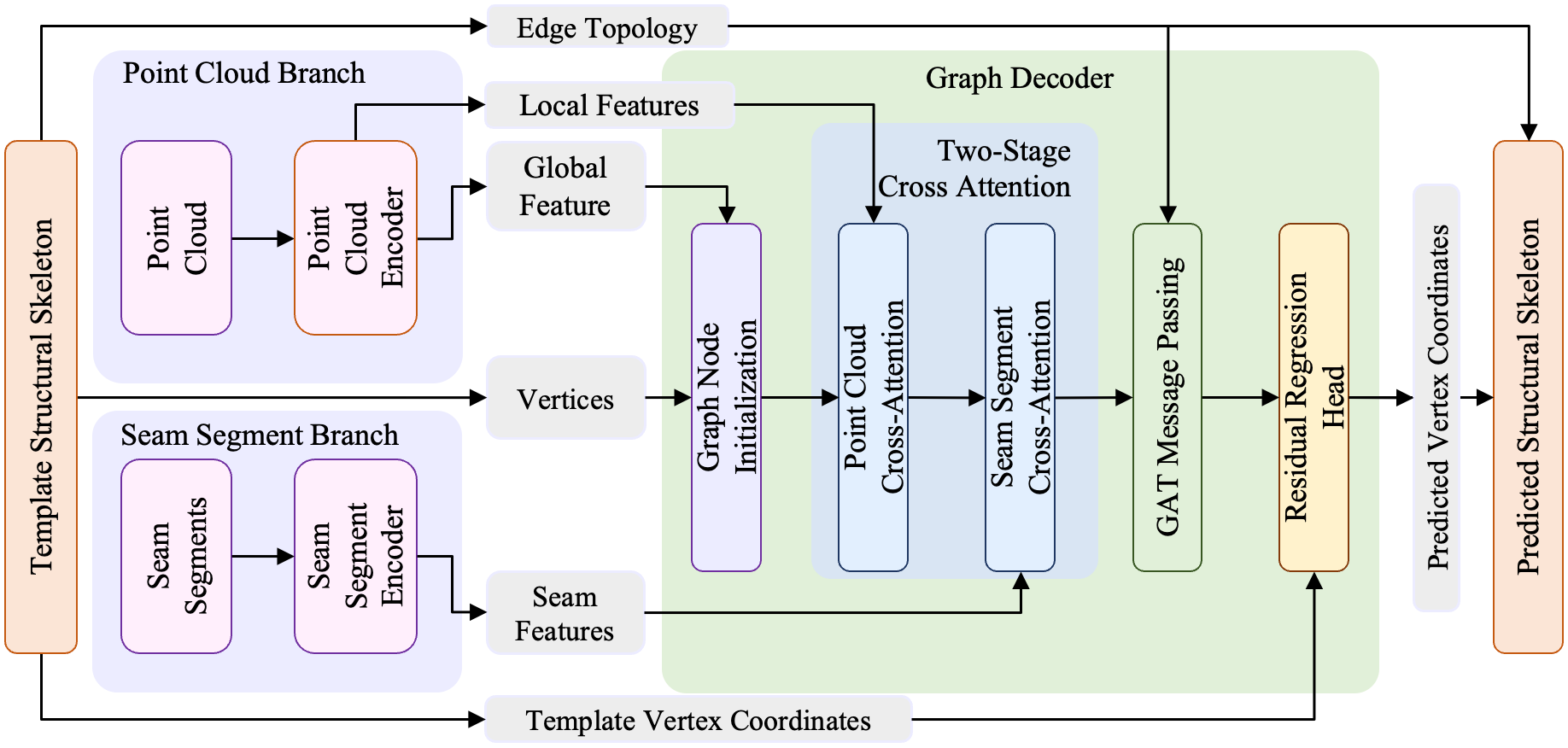}
    \caption{
        The architecture of the Seam-to-Graph network. 
        The network is modified based on a Graph Attention Network~(GAT). 
        Dual-branch encoders handle $\mathcal{P}$ and $\mathcal{S}$.
        Graph node features are initialized from a template skeleton and a global point cloud feature. 
        Cross-attention fuses the features encoded from the point cloud and the seam segments into the node features.
        With GAT message passing and a residual regression head, the current structural skeleton is output. 
    }
    \label{fig:set2graph}
\end{figure}

\subsection{Seam-to-Graph Skeleton Reconstruction}
\label{sec:seam2graph}

As shown in Fig.~\ref{fig:pipeline}(b), given the preprocessed observations $\mathcal{P}$ and $\mathcal{S}$, we describe the core component of the perception mapping $\mathcal{F}$ proposed in Eq.~\ref{equ:perception}.
The primary challenge is the \emph{Seam-to-Graph} transformation, which involves mapping these unordered, variable-length observations to a graph with fixed-topology.
To address this challenge, we build our module on a Graph Neural Network (GNN)~\cite{gnn} and attention mechanisms~\cite{attention, gat}.

As illustrated in Fig.~\ref{fig:set2graph}, the overall pipeline is as follows:
First, dedicated encoders extract feature embeddings from the seam segments and the point cloud.
The output graph is initialized using a template skeleton $\mathcal{G}_{\text{temp}}$, leveraging the graph-processing nature of GNNs to ensure an invariant topology.
Attention mechanisms then aggregate the encoded features onto these initialized graph nodes.
Next, message passing with graph attention propagates information along the graph edges to refine the node representations based on local topology.
Finally, a residual regression head uses these refined features to output the updated 3D coordinates for each vertex.
The following paragraphs will detail the specific design of our proposed Seam-to-Graph network architecture.

\subsubsection{Dual-Branch Observation Encoding}
\label{sec:dualbranch}
\blacktext{
    A dual-branch encoder extracts features for each modality to process the heterogeneous inputs:
    \begin{itemize}
        \item \textbf{Point Cloud Branch:} We use a PointNet-style encoder~\cite{pointnet} to extract a global feature \(\mathbf{F}_{\text{pcd}}^\text{G}\) of the point cloud input, as well as local point features \(\mathbf{F}_{\text{pcd}}^\text{L}\), which preserve local geometric information.
        \item \textbf{Seam Segment Branch:} For each 3D seam segment, we explicitly encode its geometric attributes, including two endpoints, the midpoint, the normalized direction, the length, and the visual category embedding as vectors. 
        A Transformer encoder then updates these vectors, allowing each segment to aggregate information from related segments. 
        Rather than compressing all segment features into a single global vector, we preserve one feature vector for each segment, denoted as \(\mathbf{F}_{\text{seg}}\), for later assignment to the graph nodes via cross-attention.
    \end{itemize}
    Notably, both \(\mathcal{P}\) and \(\mathcal{S}\) are jointly centered and normalized to guarantee translation and scale invariance.
}

\subsubsection{Attention-Guided Graph Grounding}
The decoder begins with template-initialized graph nodes and progressively grounds them to the observations through two cross-attention stages, followed by Graph Attention Network~(GAT) message passing.

\textbf{Graph Node Initialization.}
Since the template skeleton $\mathcal{G}_{\text{temp}}$ only describes a fixed, canonical configuration, it does not reflect the current garment deformation. 
Therefore, for each node $i$, we concatenate its template coordinates $\mathbf{x}_{\text{temp},i}$ with the global feature $\mathbf{F}_{\text{pcd}}^\text{G}$ extracted by the point cloud encoder. 
The concatenated vector is then passed through a Multi-Layer Perceptron (MLP) to obtain the initial feature of node $i$:
\begin{equation}
    \mathbf{f}_{\text{node}_i}^{(0)} = \operatorname{MLP}_{\text{init}}\bigl( [\mathbf{x}_{\text{temp},i},\; \mathbf{F}_{\text{pcd}}^\text{G}] \bigr),
\end{equation}
where $[\cdot, \cdot]$ denotes concatenation. 
This yields the initial node features $\mathbf{F}_{\text{node}}^{(0)}$.

\textbf{Multi-Head Attention Mechanism (Brief Overview).}
For clarity, we briefly review the attention mechanism used in the following cross-attention modules.
Given a query matrix \(\mathbf{Q}\), a key matrix \(\mathbf{K}\), and a value matrix \(\mathbf{V}\), the output is:
\begin{equation}
    \operatorname{Attn}(\mathbf{Q},\mathbf{K},\mathbf{V})
    =
    \operatorname{softmax}
    \left(
    \frac{\mathbf{Q}\mathbf{K}^{\top}}{\sqrt{d_{\mathbf{K}}}}
    \right)
    \mathbf{V},
\end{equation}
where $d_{\mathbf{K}}$ denotes the dimension of \(\mathbf{K}\).
The softmax operation produces attention weights that measure the relevance between each query and all keys. The output is a weighted aggregation of the corresponding values.

Multi-Head Attention (MHA) extends this operation by applying multiple attention heads in parallel and concatenating their outputs.
In the following cross-attention modules, we use $\operatorname{MHA}(\mathbf{Q},\mathbf{K},\mathbf{V})$ to denote MHA. 

\textbf{Point Cloud Cross-Attention.}
The initialized node features attend to the local features extracted from the point cloud.
A residual connection is then applied, followed by layer normalization denoted by $\operatorname{LN}(\cdot)$, to stabilize training.
Specifically, \(\mathbf{F}_\text{node}^{(0)}\) is used as the query, while \(\mathbf{F}_{\text{pcd}}^\text{L}\) is used as the key and value:
\begin{equation}
    \mathbf{F}_\text{node}^{(1)}
    =
    \operatorname{LN}\!\left(
    \mathbf{F}_\text{node}^{(0)}
    +
    \operatorname{MHA}
    \bigl(
    \mathbf{Q}=\mathbf{F}_\text{node}^{(0)},
    \mathbf{K}=\mathbf{V}=\mathbf{F}_{\text{pcd}}^\text{L}
    \bigr)
    \right).
\end{equation}

At this stage, each template node is enriched with local geometric cues from relevant regions of the point cloud. 
Rather than directly changing the coordinates, it updates node features. 

\textbf{Seam Segment Cross-Attention.}
The node features then attend to the seam segment features. 
We use \(\mathbf{F}_\text{node}^{(1)}\) as the query, and seam segment features \(\mathbf{F}_{\text{seg}}\) as the key and value:
\begin{equation}
    \mathbf{F}_\text{node}^{(2)}
    =
    \operatorname{LN}\!\left(
    \mathbf{F}_\text{node}^{(1)}
    +
    \operatorname{MHA}
    \bigl(
    \mathbf{Q}=\mathbf{F}_\text{node}^{(1)},
    \mathbf{K}=\mathbf{V}=\mathbf{F}_{\text{seg}}
    \bigr)
    \right).
\end{equation}
Because each segment feature contains geometric information and a visual category label, attention weights can learn node-segment correspondences without the need for hand-crafted matching rules. 
The residual connection preserves the template-derived node context while introducing seam cues.

\textbf{Topology-Aware Message Passing.}
The cross-attention stages provide spatial and semantic evidence, but local inconsistencies among adjacent nodes may still remain due to observation noise. 
Therefore, we apply a Graph Attention Network~(GAT)~\cite{gat} along the template edges $\mathcal{E}$ to aggregate information from direct topological neighbors, yielding the final refined node features $\mathbf{F}_\text{node}$.

\subsubsection{Residual Regression Head}
A regression head, which is a small MLP, maps the refined node features$\mathbf{F}_\text{node}$ to per-vertex residual displacements:
\begin{equation}
    \tilde{\mathbf{X}} = \mathbf{X}_{\text{temp}} + \operatorname{MLP}_\text{final}(\mathbf{F}_\text{node}).
\end{equation}
This residual formulation restricts the learning problem to predicting deformations from the template skeleton, thereby stabilizing the training process. 
The predicted skeleton $\tilde{\mathcal{G}}=\{\mathcal{V}, \mathcal{E}, \tilde{\mathbf{X}}\}$ inherits the topological semantics of the template skeleton because they share the same edge connectivity $\mathcal{E}$.

\subsubsection{Loss Function}
The network is trained using a loss function:
\begin{equation}
    \begin{aligned}
        \mathcal{L}_{\text{train}} = & \ \lambda_{\text{vtx}} \underbrace{\frac{1}{N_{\text{vtx}}}\sum_{i=1}^{N_{\text{vtx}}}\|\tilde{\mathbf{X}}_i - \mathbf{X}_{\text{gt},i}\|_2^2}_{\mathcal{L}_{\text{vtx}}} \\
        +                            & \ \lambda_{\text{edge}} \underbrace{\frac{1}{|\mathcal{E}|}\sum_{(i,j)\in\mathcal{E}} \bigl| \|\tilde{\mathbf{X}}_i-\tilde{\mathbf{X}}_j\| - \|\mathbf{X}_{\text{gt},i}-\mathbf{X}_{\text{gt},j}\| \bigr|}_{\mathcal{L}_{\text{edge}}},
    \end{aligned}
\end{equation}
where $\lambda_{\text{vtx}}$ and $\lambda_{\text{edge}}$ are weights. 
$\mathcal{L}_{\text{vtx}}$ is the mean squared error~(MSE) on vertex positions, and $\mathcal{L}_{\text{edge}}$ penalizes deviations in edge lengths to preserve the global shape.

\subsection{Topology-Constrained Skeleton Optimization}
\label{subsec:optimization}

Although the network provides a robust raw prediction $\tilde{\mathbf{X}}$, geometric misalignments may persist between the raw prediction and the physical observations.
As shown in Fig.~\ref{fig:pipeline}(c), to obtain the final estimated state $\hat{\mathcal{G}} = (\mathcal{V}, \mathcal{E}, \hat{\mathbf{X}})$ required by the controller in Eq.~\ref{equ:control}, we refine the predicted vertex positions by minimizing a composite energy function:
\begin{equation}
    \begin{split}
        \min_{\mathbf{X}} E(\mathbf{X}) & = \frac{\lambda_{\text{seg}}}{2N_\text{seg}} \sum_{k=1}^{2 N_\text{seg}} \text{D}_{\mathbf{X}}^2(\mathbf{p}^\text{end}_{k}) + \frac{\lambda_{\text{pcd}}}{|\Omega|} \sum_{\mathbf{p} \in \Omega} \text{D}_{\mathbf{X}}^2(\mathbf{p}) \\
                                        & \quad + \frac{\lambda_{\text{len}}}{|\mathcal{E}|} \sum_{(i,j) \in \mathcal{E}} \big| \|\mathbf{X}_i - \mathbf{X}_j\| - \|\tilde{\mathbf{X}}_i - \tilde{\mathbf{X}}_j\| \big|,
    \end{split}
    \label{equ:optimization}
\end{equation}
where $\lambda_{\text{seg}}, \lambda_{\text{pcd}}, \lambda_{\text{len}}$ are weights. 
$\mathbf{p}$ denotes a point in 3D space, and $\mathbf{p}^\text{end}_{k}$ denotes the $k^\text{th}$ endpoint extracted from the seam segments $\mathcal{S}$.
The function \(\text{D}_{\mathbf{X}}(\mathbf{p}) = \min_{(i,j)\in\mathcal{E}} \operatorname{dist}(\mathbf{p}, \overline{\mathbf{X}_i\mathbf{X}_j})\) denotes the Euclidean distance from a point to its nearest skeleton edge.
The first term pulls the predicted skeleton edges toward the endpoints of the detected seam segments.
The second term attracts the predicted skeleton edges toward a set of nearby surface points $\Omega \subset \mathcal{P}$ defined as $\Omega = \{\mathbf{p} \in \mathcal{P} \mid \text{D}(\mathbf{p}) < \delta\}$.
The set $\Omega$ is selected before optimization using the raw prediction $\tilde{\mathbf{X}}$.
The third term penalizes deviations in edge lengths relative to the raw prediction $\tilde{\mathbf{X}}$, preserving the learned deformation.
This nonlinear problem is efficiently solved via gradient descent using the Adam optimizer~\cite{adam} with cosine annealing, yielding the refined skeleton vertex coordinates $\hat{\mathbf{X}}$.

\begin{figure}[tbp]
    \includegraphics[width=\linewidth]{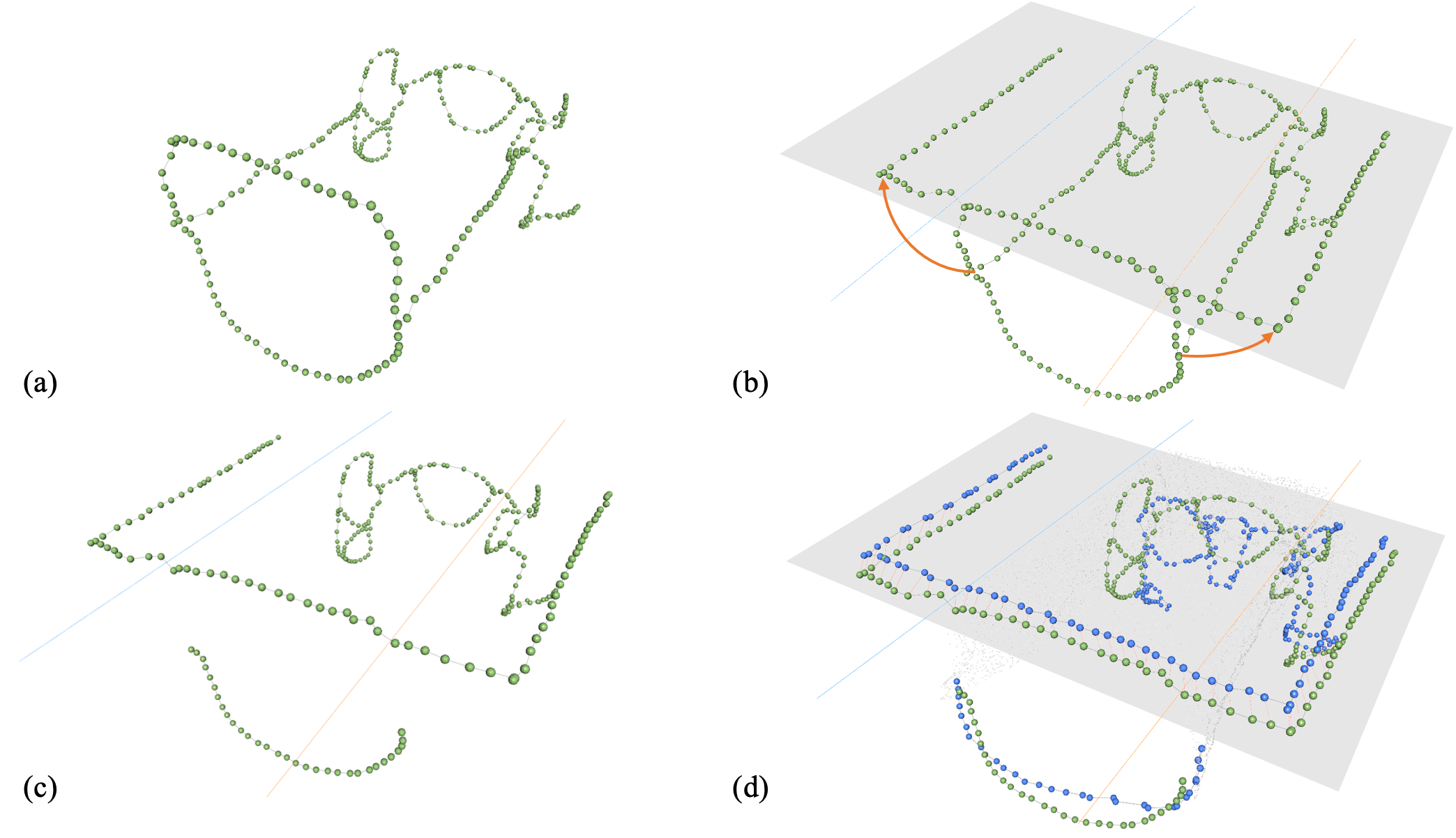}
    \caption{
        Unfolding operation to regional skeleton vertices. 
        (a)(b)(c) A geometric mapping process is applied to unfold the estimated vertices under the platen onto the platen plane.
        (d) The current estimated skeleton (blue vertices) and the desired skeleton (green vertices) undergo this unfolding operation, which formulates the control problem as a planar $\mathbb{SE}(2)$ problem.
    }
    \label{fig:unfolding}
\end{figure}

After this refinement, the current skeleton \(\hat{\mathcal{G}} = (\mathcal{V}, \mathcal{E}, \hat{\mathbf{X}})\) is obtained from real-time observations as shown in Fig.~\ref{fig:unfolding}(a).
The desired skeleton \(\mathcal{G}^{\star}\) is constructed from a human-demonstrated aligned configuration using the same pipeline.

\subsection{Deformation-Aware Visual Servoing Controller}

Based on the estimated skeleton \(\hat{\mathcal{G}}\) and the goal skeleton \(\mathcal{G}^{\star}\), we address the control subproblem in Eq.~\ref{equ:control} (see Fig.~\ref{fig:pipeline}(d)). 
Since a deformable garment cannot be represented by a single global rigid pose, we decompose the alignment task into a set of locally quasi-rigid regions on the skeleton.

The proposed controller consists of four components:
\begin{itemize}
    \item \textbf{Unfolding}: Each skeleton region is unfolded onto the platen plane, converting the deformation-aware alignment problem into a set of 2D planar registration tasks.
    \item \textbf{Regional pose error estimation}: The 2D Kabsch algorithm~\cite{kabsch1978discussion} registers the unfolded current skeleton region to its unfolded desired counterpart, yielding a planar pose error.
    \item \textbf{Inner PBVS loop}: The estimated regional pose error is converted into incremental bimanual end-effector displacements via a position-based visual servoing (PBVS) controller.
    \item \textbf{Outer coordination loop}: A region-level coordinator iteratively selects regions that are out of tolerance and invokes the inner PBVS loop, until all regional errors satisfy the prescribed tolerances.
\end{itemize}

\subsubsection{Unfolding and Local Quasi-Rigidity}
To make the deformable garment alignment problem tractable under the platen support constraint, we first partition the skeleton into \(R\) local regions \(\{\mathcal{V}_r\}_{r=1}^{R}\).
For each region, we apply an unfolding operation that maps the 3D vertices onto the platen plane, which is parallel to the \(xy\)-plane of \(\Sigma\)). 
Vertices below the platen surface are rotated about the platen edge onto the plane, and those above the platen plane are projected onto it (see Fig.~\ref{fig:unfolding}).
This yields a 2D representation of each region on the platen plane. 
The desired skeleton is also represented in this plane.

On this unfolded 2D plane, we assume local quasi-rigidity, meaning garment motion is dominated by in-plane translation and rotation within each region and over the course of one PBVS correction step. 
This assumption is motivated by the gravitational tension of the hanging garment portions. 
It allows us to approximate the regional alignment error as a planar \(\mathbb{SE}(2)\) transformation rather than full 3D deformable motion. 
Consequently, the alignment problem can be solved using an \(\mathbb{SE}(2)\) registration.
Any residual non-rigid motion that violates the quasi-rigid approximation is corrected by subsequent visual-feedback iterations.

\subsubsection{Regional Planar Pose Error Calculation}
In the unfolding-and-local-rigidity formulation above, the pose error of region \(r\) is defined as the corrective planar $\mathbb{SE}(2)$ transformation that best maps the current unfolded skeleton vertices to their desired counterparts.
We estimate this transformation using the 2D Kabsch algorithm.
Let \(\hat{\mathbf{P}}_r, \mathbf{P}^{\star}_r \in \mathbb{R}^{|\mathcal{V}_r| \times 2}\) be the unfolded current and desired 2D coordinates of region \(r\), and let \(\hat{\mathbf{p}}_{\mathrm{cent},r}, \mathbf{p}^{\star}_{\mathrm{cent},r} \in \mathbb{R}^{2}\) denote their respective centroids.
Because the estimated and desired skeletons share the same topology, point-wise correspondences are available directly.
The cross-covariance matrix is constructed as:
\begin{equation}
    \mathbf{H}_r =
    \left(\hat{\mathbf{P}}_r - \mathbf{1}\hat{\mathbf{p}}_{\mathrm{cent},r}^{\mathsf{T}}\right)^{\mathsf{T}}
    \left(\mathbf{P}^{\star}_r - \mathbf{1}{\mathbf{p}^{\star}_{\mathrm{cent},r}}^{\mathsf{T}}\right),
    \label{equ:covariance_2d}
\end{equation}
where \(\mathbf{1} \in \mathbb{R}^{|\mathcal{V}_r|}\) is an all-ones vector.
After decomposing \(\mathbf{H}_r = \mathbf{U}\boldsymbol{\Sigma}\mathbf{V}^{\mathsf{T}}\) and applying the Kabsch correction \(\mathbf{S} = \operatorname{diag}(1, \det(\mathbf{V}\mathbf{U}^{\mathsf{T}}))\) to prevent reflections~\cite{kabsch1978discussion}, the optimal corrective rotation \(\mathbf{R}^{*}_r\) and translation \(\mathbf{t}_r = [t_{x,r}, t_{y,r}]^{\mathsf{T}}\) are given by:
\begin{equation}
    \mathbf{R}^{*}_r = \mathbf{V}\mathbf{S}\mathbf{U}^{\mathsf{T}}, \quad
    \mathbf{t}_r = \mathbf{p}^{\star}_{\mathrm{cent},r} - \mathbf{R}^{*}_r \hat{\mathbf{p}}_{\mathrm{cent},r}.
    \label{equ:local_rt}
\end{equation}
The in-plane rotation error is extracted as \(\theta_{z,r} = \operatorname{atan2}(\mathbf{R}^{*}_{r,21}, \mathbf{R}^{*}_{r,11})\).
This yields the regional planar pose error \(\mathbf{e}_r = [t_{x,r}, t_{y,r}, \theta_{z,r}]^{\mathsf{T}}\), which is recomputed from the latest visual observations during visual servoing.

\subsubsection{Inner-Loop Regional PBVS}
Given the regional pose error \(\mathbf{e}_r\), the inner loop executes incremental PBVS corrections for the selected region until its error falls within the prescribed tolerance.
For notational clarity, this subsection omits the region index \(r\) and write \(\mathbf{e} = [t_x, t_y, \theta_z]^{\mathsf{T}}\) for the currently selected region.
At each servo iteration, the garment skeleton is updated using the latest observations. Then, the regional pose error is recomputed, and an incremental bimanual command is generated.
This feedback design prevents the controller from relying on an open-loop, rigid-motion prediction of the deformable garment.

A virtual control point $\mathbf{p}_\text{ctrl} \in \mathbb{R}^2$ is defined as the midpoint of the two grasping points projected onto the $xy$-plane.
Since bimanual motion induces rotation around \(\mathbf{p}_{\mathrm{ctrl}}\) rather than the world-frame origin, the translation component is compensated as:
\begin{equation}
    \begin{bmatrix}
        t_x' \\
        t_y'
    \end{bmatrix}
    =
    \begin{bmatrix}
        t_x \\
        t_y
    \end{bmatrix}
    -
    (\mathbf{I}_2 - \mathbf{R}^{*})\mathbf{p}_{\mathrm{ctrl}},
    \label{equ:control_point_compensation}
\end{equation}
where \(\mathbf{I}_2 \in \mathbb{R}^{2 \times 2}\) is the identity matrix.
The compensated error is denoted as \(\mathbf{e}' = [t_x', t_y', \theta_z]^{\mathsf{T}}\).

To obtain smooth incremental commands and reduce overshoot, we scale the compensated error using a nonlinear diagonal adaptive gain:
\begin{equation}
    \begin{bmatrix}
        \Delta \mathbf{t}_{\mathrm{ctrl}} \\
        \Delta \theta_z
    \end{bmatrix}
    =
    \left(
    k_{\min}\mathbf{I}_3
    +
    (k_{\max} - k_{\min})
    \operatorname{diag}
    \left(
        \frac{\boldsymbol{\epsilon}}{|\mathbf{e}'| + \boldsymbol{\epsilon}}
    \right)
    \right)\mathbf{e}'.
    \label{equ:scaled_error}
\end{equation}
where \(\Delta \mathbf{t}_{\mathrm{ctrl}} = [\Delta t_x, \Delta t_y]^{\mathsf{T}}\), \(\mathbf{I}_3 \in \mathbb{R}^{3 \times 3}\), \(|\cdot|\) denotes the element-wise absolute value, and \(\boldsymbol{\epsilon} = [\epsilon_x, \epsilon_y, \epsilon_{\theta}]^{\mathsf{T}}\) is a small positive vector that controls the transition sensitivity between $k_{\min}$ and $k_{\max}$.
The gain approaches \(k_{\min}\) for large errors and \(k_{\max}\) for small residual errors.

\begin{algorithm}[t]
    \caption{Deformation-Aware Visual Servoing Controller}
    \label{alg:servo_loop}
    \begin{algorithmic}
        \Require Desired skeleton $\mathcal{G}^{\star}$, template skeleton $\mathcal{G}_{\text{temp}}$
        \State AllRegionsWithinTolerance \(\gets\) \textbf{false}
        \While{$\neg$ \text{AllRegionsWithinTolerance}}
        \State $\mathcal{P}, \mathcal{S} \gets \text{updateObservations}()$
        \State $\hat{\mathcal{G}} \gets \text{SkeletonEstimation}(\mathcal{P}, \mathcal{S}, \mathcal{G}_{\text{temp}})$
        \State AllRegionsWithinTolerance $\gets$ \textbf{true}
        \For{each region $r$}
        \State $\mathbf{e}_r \gets \text{EstimateRegionalPoseError}(r, \hat{\mathcal{G}}, \mathcal{G}^{\star})$
        \If{$|\mathbf{e}_r[j]| > e^{\text{tolerance}}[j]$ for any $j$}
        \State AllRegionsWithinTolerance $\gets$ \textbf{false}
        \State $\text{ExecuteRegionalPBVS}(\mathbf{e}_r, r, \mathcal{G}^{\star}, \mathcal{G}_{\text{temp}})$
        \State \textbf{break} \Comment{Execute one region's move at a time}
        \EndIf
        \EndFor
        \EndWhile
        \vspace{0.5em}
        \Statex \textbf{Function} $\text{ExecuteRegionalPBVS}(\mathbf{e}_r, r, \mathcal{G}^{\star}, \mathcal{G}_{\text{temp}})$:
        \While {$\exists j$ such that $|\mathbf{e}_r[j]| > e^{\text{tolerance}}[j]$}
        \State $\mathbf{p}_{\text{ctrl}} \gets \text{UpdateCtrlPointFromRobotState}()$
        \State $\mathbf{e}'_r \gets \text{CtrlPointCompensation}(\mathbf{e}_r, \mathbf{p}_{\text{ctrl}})$ \Comment{Eq.~\ref{equ:control_point_compensation}}
        \State $\Delta \mathbf{t}_{\text{ctrl}}, \Delta\theta_z \gets \mathbf{K}(\mathbf{e}'_r) \cdot \mathbf{e}'_r$ \Comment{Eq.~\ref{equ:scaled_error}}
        \For{each end-effector \(a \in \{\text{L}, \text{R}\}\)}
        \State \(\Delta \mathbf{t}_{a} \gets \text{GetArmCmd}(\Delta \mathbf{t}_{\text{ctrl}}, \Delta\theta_z, a)\)
        \EndFor
        \State \(\text{SendBimanualCmd}(\{\Delta \mathbf{t}_{a}\}_{a \in \{\text{L}, \text{R}\}})\)
        \State \(\mathcal{P}, \mathcal{S} \gets \text{updateObservations}()\)
        \State \(\hat{\mathcal{G}} \gets \text{SkeletonEstimation}(\mathcal{P}, \mathcal{S}, \mathcal{G}_{\text{temp}})\)
        \State \(\mathbf{e}_r \gets \text{EstimateRegionalPoseError}(r, \hat{\mathcal{G}}, \mathcal{G}^{\star})\)
        \EndWhile
    \end{algorithmic}
\end{algorithm}

\begin{figure}[t]
    \includegraphics[width=\linewidth]{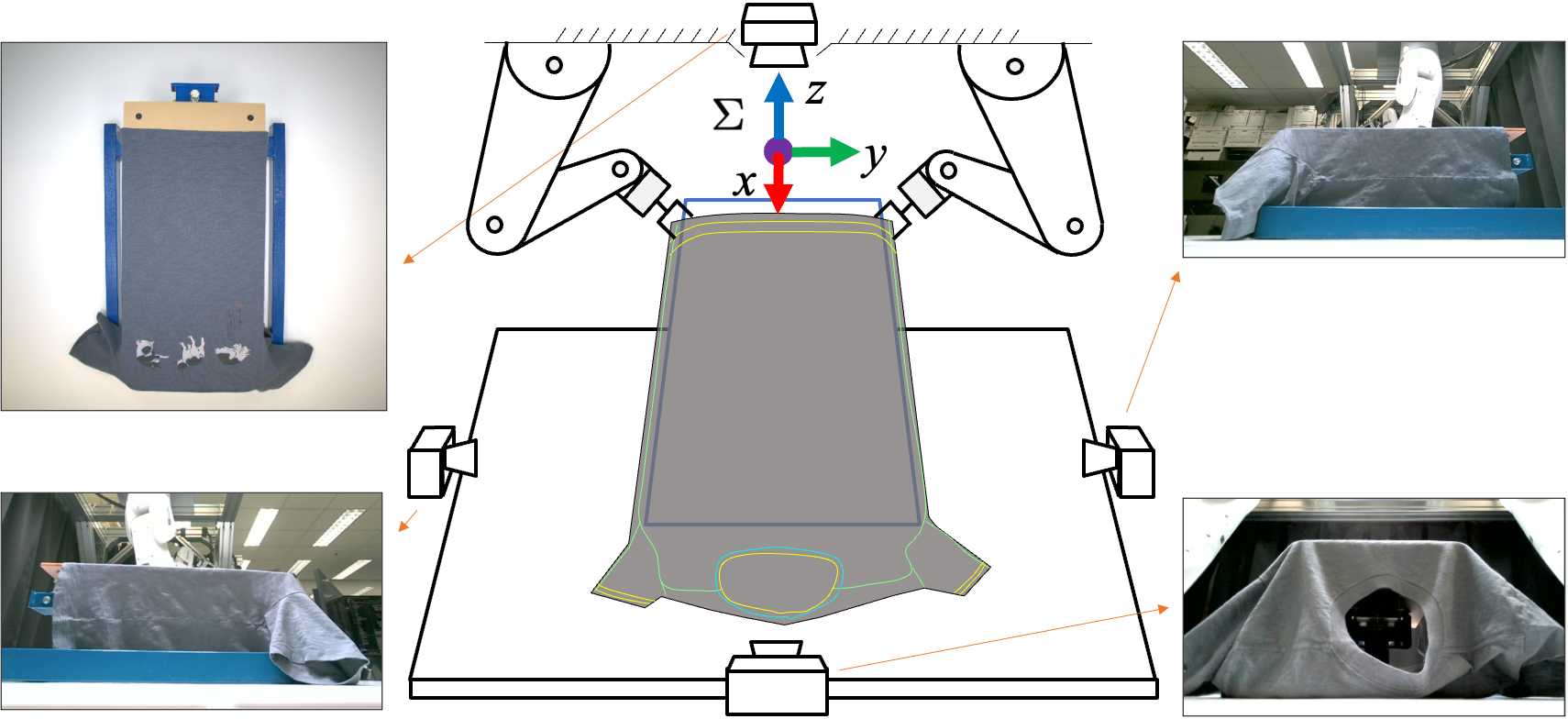}
    \caption{
        Hardware platform for experiments. 
        A screen printing platen is attached to the table. 
        The platen plane is parallel to the $xy$-plane of the world frame $\Sigma$.
        Four RGBD cameras are installed to capture multi-view observations of the garment. 
    }
    \label{fig:platform}
\end{figure}

Assuming that the two grasping points follow the commanded planar rigid motion around \(\mathbf{p}_{\mathrm{ctrl}}\), the end-effector displacement of arm \(a\) is computed as:
\begin{equation}
    \Delta \mathbf{t}_a
    =
    \Delta \mathbf{t}_{\mathrm{ctrl}}
    +
    \left(\mathbf{R}_z(\Delta\theta_z) - \mathbf{I}_2\right)
    \left(\mathbf{p}_a - \mathbf{p}_{\mathrm{ctrl}}\right),
    \label{equ:ee_mapping}
\end{equation}
where \(\Delta \mathbf{t}_a = [\Delta t_{x,a}, \Delta t_{y,a}]^{\mathsf{T}}\), \(\mathbf{p}_a\) is the projected end-effector position of arm \(a\), \(a \in \{\mathrm{L}, \mathrm{R}\}\) (i.e. left and right arms), and \(\mathbf{R}_z(\Delta\theta_z)\) is the 2D rotation matrix associated with the incremental rotation \(\Delta\theta_z\).

These synchronized bimanual displacements are executed through the underlying motion controller.
Specifically, fifth-order polynomial interpolation generates smooth Cartesian trajectories, impedance control regulates contact force, and inverse kinematics converts Cartesian commands into joint commands \(\mathbf{q}_{\mathrm{cmd}}\).

\subsubsection{Outer-Loop Region-level Coordination}
The outer loop addresses the inter-region coupling caused by the garment deformability.
Since the garment is not rigid, correcting one region can affect previously corrected regions.
Thus, the inner regional PBVS loop is embedded in an outer supervisory loop that repeatedly checks and corrects regions that are out-of-tolerance.
Algorithm~\ref{alg:servo_loop} summarizes the coordination between the outer loop and the inner-loop regional PBVS.
This scan-and-correct process continues until all regional errors are within tolerance, resulting in whole-garment alignment on the platen with the specified level of accuracy.

\begin{figure*}[t]
    \includegraphics[width=\linewidth]{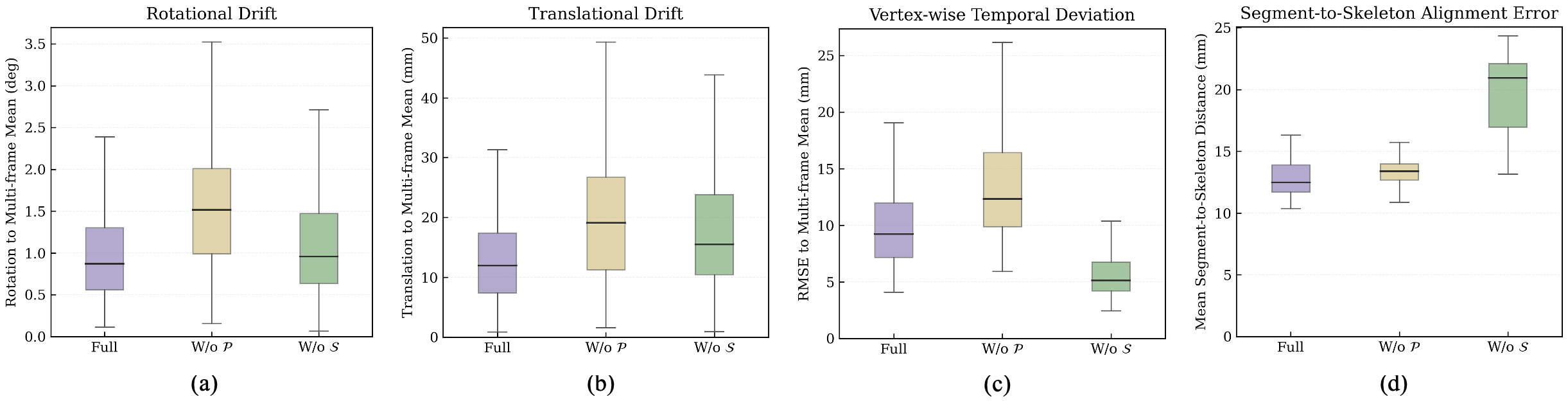}
    \caption{
        Ablation study on real-world repeatability and segment-to-skeleton alignment. 
        The Full group uses our proposed complete method with a dual-branch architecture. 
        The W/o \(\mathcal{P}\) and W/o \(\mathcal{S}\) remove the point cloud and the seam segment branches, respectively. 
        We report four metrics: 
        (a) rotational drift and 
        (b) translational drift estimated using the Kabsch algorithm; 
        (c) vertex-wise RMSE with respect to the multi-frame mean skeleton; 
        and (d) the mean distance from the detected seam-segment endpoints to the estimated skeleton edges. 
        Lower values indicate better repeatability or better alignment accuracy.
    }
    \label{fig:repeatability}
\end{figure*}

\section{Experiments}\label{sec:experiment}

This section presents a comprehensive evaluation of the proposed scheme using screen printing as an example application.
First, we validate the accuracy and repeatability of the perception module using an unseen synthetic test set and real-world continuous sampling.
Then, we conduct extensive robotic experiments to demonstrate garment alignment accuracy, generalization across different garments, and robustness to initial grasping deviations.

\subsection{Experimental Setup and Training}

\textbf{Hardware Platform.}
The platform simulates the scenario of automatically loading garments onto a printing platen (see Fig.~\ref{fig:platform}).
A rigid screen printing platen (580 mm $\times$ 280 mm) is mounted on the table.
The robot system consists of two fixed Denso VS087 robot arms, each of which is equipped with an ATI Axia80-M8 force torque sensor and a Taiyo EGS2-LS-4230 parallel gripper.
Four RGB-D cameras (two Intel RealSense D455, one D435i, and one Azure Kinect DK) provide multi-view observation inputs.
In the evaluation phase, a downward-facing, high-resolution Basler acA4096-30uc camera is used to obtain final pattern poses. This camera achieves $0.2$ mm/pixel measurement accuracy when equipped with an 8 mm Basler lens.
The system runs on a PC with an NVIDIA RTX 5090 GPU and an Intel i9-14900 CPU, and uses ROS2~\cite{ros2}.

\begin{table}[t]
    \centering
    \caption{Perception ablation on the unseen test set.}
    \label{tab:perception_ablation}
    \small
    \setlength{\tabcolsep}{4.0pt}
    \begin{tabular}{lccccc}
        \hline
        \textbf{Group}             & $\mathcal{P}$ & $\mathcal{S}$ & \textbf{Vtx Loss}   & \textbf{Edge Loss}  & \textbf{MV2E Dist}\\
                                   &               &               & $\mathcal{L}_{\text{vtx}}$ ($\times 10^{-2}$) & $\mathcal{L}_{\text{edge}}$ ($\times 10^{-2}$) & ($\times 10^{-2}$)\\
        \hline
        \textbf{W/o $\mathcal{S}$} & $\checkmark$  & $\times$      & 0.6168               & 0.8713 & 3.9506             \\
        \textbf{W/o $\mathcal{P}$} & $\times$      & $\checkmark$  & 0.3750               & \textbf{0.8217} & 2.8437             \\
        \textbf{Full}       & $\checkmark$  & $\checkmark$  & \textbf{0.3357}      & 0.8502     & \textbf{2.6854} \\
        \hline
    \end{tabular}
\end{table}

\textbf{Simulation and Training.}
The network is trained on 99,000 synthetic samples generated via Blender using a T-shirt mesh model from~\cite{garmentlibrary,garmentlibrary2}.
To bridge the sim-to-real gap, we apply domain randomization to $\mathcal{P}$ (viewpoint clipping, occluders) and $\mathcal{S}$ (segment dropping, endpoint jitter, category label flipping).
We optimize the model using AdamW with an initial learning rate of 0.001 and a cosine annealing schedule over 600 epochs.
The loss weights are $\lambda_{\text{vtx}} = 1.0$ and $\lambda_{\text{edge}} = 0.1$.
The prediction frequency of the Seam-to-Graph network is 9.9~Hz.

The post-processing optimization in Sec.~\ref{subsec:optimization} is iterated 50 times with an initial learning rate of 0.01.
The learning rate decays following a cosine annealing schedule.
The energy function weights are $\lambda_{\text{seg}} = 0.7$, $\lambda_{\text{pcd}} = 0.2$, and $\lambda_{\text{len}} = 0.1$.

\subsection{Perception Module Evaluation}

\subsubsection{Network Ablation on Unseen Synthetic Test Set}
We evaluate the generalization of the Seam-to-Graph network using an unseen, synthetic test set of 11,000 samples with novel garment configurations.
To verify the necessity of the dual-branch architecture in Sec.~\ref{sec:seam2graph}, we conduct ablation studies against our proposed method (the Full group) by independently disabling the point cloud branch and the seam segment branch.

Table~\ref{tab:perception_ablation} presents the quantitative results.
The unit of the metrics is normalized distance in the simulator. 
Removing the seam segment branch (the W/o $\mathcal{S}$ group) significantly increases both the edge loss $\mathcal{L}_{\text{edge}}$ and the vertex loss $\mathcal{L}_{\text{vtx}}$ compared to the full method.
This indicates that seams provide critical structural information and absolute anchoring for accurate garment configuration estimation.
Conversely, disabling the point cloud branch (the W/o $\mathcal{P}$ group) degrades the vertex loss $\mathcal{L}_{\text{vtx}}$, confirming its role in providing geometric constraints for skeleton vertices and complementing the seam segment branch, particularly in regions where seam segments are misdetected.
Additionally, we measure the mean distance from the predicted vertices to the ground truth skeleton edges (MV2E Dist) for these groups. 
The Full group outperforms both ablation groups, demonstrating that its predicted skeleton is closer to the ground truth. 
These experimental results demonstrate that $\mathcal{S}$ provides absolute anchoring through structural cues, while $\mathcal{P}$ provides geometric constraints, enabling the full model to achieve precise localization.

\subsubsection{Temporal Repeatability and Accuracy Analysis of Real-World Estimation}
We evaluate the real-world performance of the proposed state estimation pipeline using data collected from a static T-shirt placed on the platen. 
The input sequence is obtained by capturing 100 frames continuously, and a structural skeleton is estimated for each frame. 
Since ground-truth skeletons are unavailable in the real world, we use the multi-frame mean skeleton as a pseudo-reference baseline. 
Specifically, we compute the mean coordinate of each corresponding vertex over all frames and measure the frame-wise RMSE with respect to this baseline. 
We then estimate the rigid transformation between each skeleton and the mean skeleton using the 3D Kabsch algorithm, and report the rotation angle and translation magnitude as temporal drift metrics. 
Finally, since temporal repeatability alone does not ensure proper alignment with local garment structures, we calculate the mean distance between the detected seam-segment endpoints and the nearest estimated skeleton edges. This distance is referred to as the segment-to-skeleton distance. 
This metric evaluates the local geometric alignment with the observed seam cues. A smaller value indicates better alignment.

\begin{figure*}[t]
    \includegraphics[width=\linewidth]{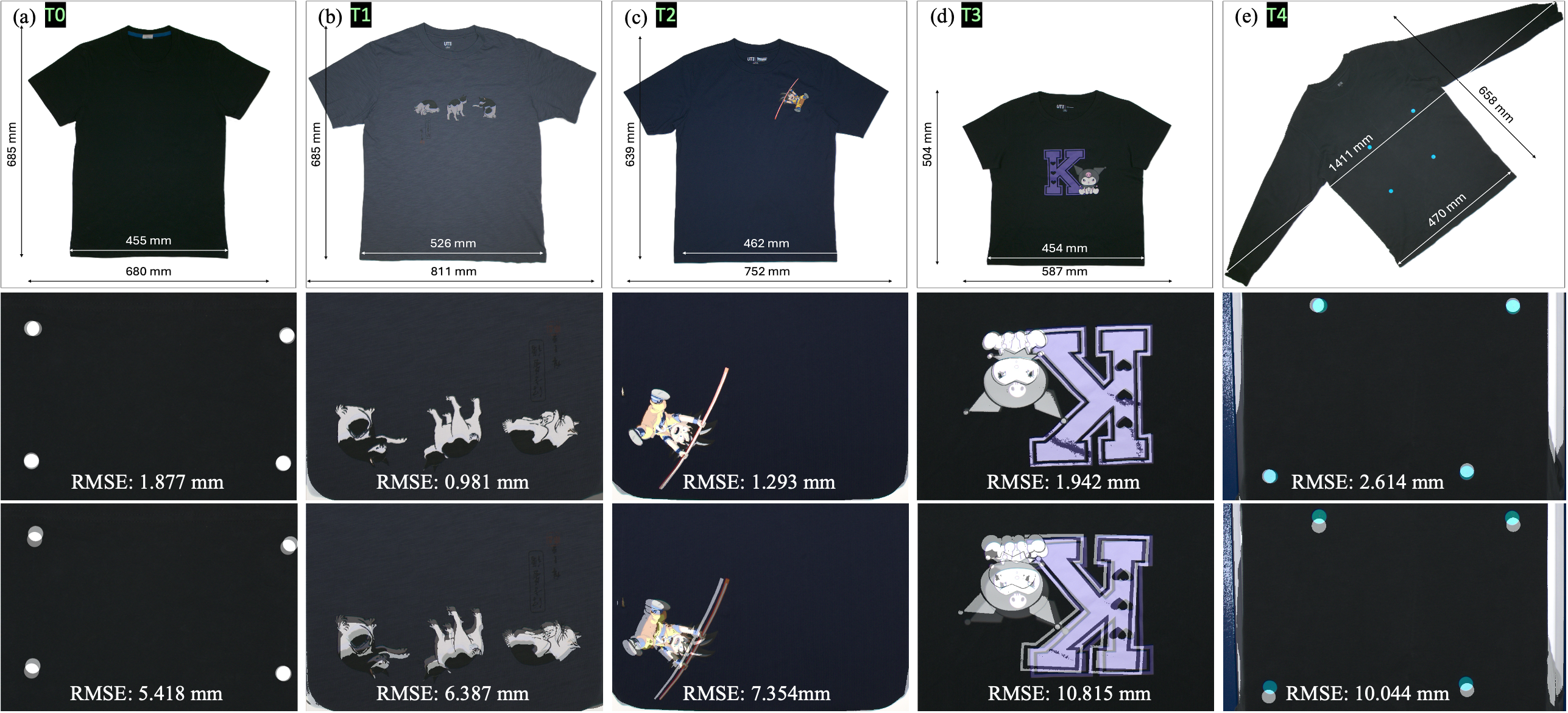}
    \caption{
        Clothing and qualitative results.
        The first row shows the garments used during the experiments. 
        Our Seam-to-Graph network is only trained on simulated data. Therefore, these garments are unseen. 
        T4 is a long-sleeved T-shirt, which differs from the mesh model used to generate the simulation dataset.
        The second and third rows present the qualitative alignment results. 
        For each sample, we superimpose an image of the aligned garment onto an image of the garment in the desired configuration to visualize the alignment error. 
        The second row shows the best result (lowest RMSE), and the third row shows the worst result (highest RMSE) over ten trials for each garment. 
        The corresponding RMSE value is marked on each image. 
        Note that the feature points overlaid on the T0 results are manually added for visualization purposes and are not present on the actual garment.
    }
    \label{fig:cloth}
\end{figure*}

\begin{figure}[t]
    \includegraphics[width=\linewidth]{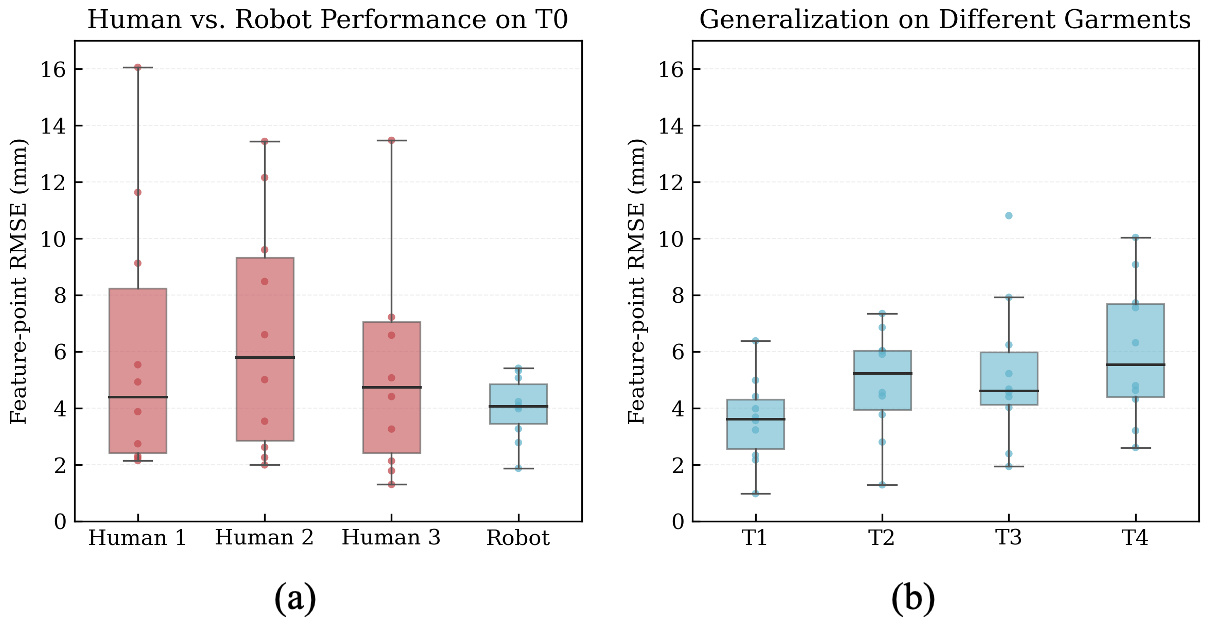}
    \caption{
        Evaluation of garment loading and alignment performance. 
        We quantify operation quality using the RMSE of four feature points on the printed pattern between the operated and desired configurations. 
        The boxplots show the results of ten trials. The red and blue boxes represent results of human operators and the robot system, respectively. 
        (a) A comparison of human operators and the robot system on an unseen T-shirt T0 (see Fig.~\ref{fig:cloth}) during the training of the Seam-to-Graph network. 
        (b) Generalization evaluation on different unseen T-shirts, including short-sleeved T1–T3 and long-sleeved T4 (see Fig.~\ref{fig:cloth}).
    }
    \label{fig:compare}
\end{figure}

As shown in Fig.~\ref{fig:repeatability}, the full model achieves the best overall balance between temporal repeatability and local geometric alignment in real-world skeleton estimation. 
In terms of temporal repeatability, the W/o \(\mathcal{P}\) group exhibits notably higher vertex-wise RMSE, rotational drift, and translational drift than the Full group, with larger error ranges. 
This indicates that the point cloud branch provides stable geometric constraints across frames and effectively reduces inter-frame jitter. 
In contrast, the W/o \(\mathcal{S}\) group has a lower vertex-wise RMSE than the Full group, but has slightly larger rigid-transformation drift and a substantially larger segment-to-skeleton distance. 
This suggests that using only point cloud information produces temporally smooth estimations in terms of vertex-wise deviation, but weakens spatial alignment between the estimated skeleton and seam structures. 
Overall, the point cloud branch improves temporal repeatability, and the seam segment branch improves the local geometric accuracy of skeleton estimation. 
Combining both types of information in the full model achieves a better trade-off between temporal repeatability and skeleton estimation accuracy, confirming the necessity and complementarity of the dual-branch architecture.

\subsection{Manipulation Performance}

We evaluate the performance of the proposed garment loading and alignment scheme using a real bimanual robot system with screen printing as the example. 
In each trial, two grippers initially grasp the garment. 
First, the robot executes a sleeving motion to load the garment onto the platen. Then, it adjusts the garment configuration using the proposed deformation-aware visual servoing controller. 
This control loop repeats until all regional errors satisfy the prescribed tolerances or the maximum of ten regional PBVS loops is reached.
We use the RMSE of the feature points in the $xy$-plane as the performance metric for garment loading and alignment in screen printing.

\subsubsection{Comparison with Human Operators}
To validate the feasibility of the proposed method for automated screen-printing preparation, we compare its alignment quality with that of human operators under a pre-printing condition in which no visible pattern reference is available. 
We use a plain, solid-color garment T0, as shown in Fig.~\ref{fig:cloth}(a), as the experimental object. 
Four evaluation feature points are drawn on the garment using transparent fluorescent ink. 
These points are not visible during manipulation and are only used for post-trial evaluation. This ensures that neither the human operators nor the robot can use them directly as visual alignment references.

To make the comparison consistent with practical industrial operations, we allow human operators to use a standard laser alignment tool for assistance. 
In contrast, the robotic system relies on the estimated structural skeleton and the proposed control scheme. 
The comparison includes three human operators and the robotic system, each of which performs ten trials.

As shown in Fig.~\ref{fig:compare}(a), the robotic system achieves a comparable or lower median RMSE than the human operators, while also showing a noticeably tighter error distribution. 
This comparison shows that the robotic system can achieve human-level alignment quality with improved repeatability under the same pre-printing conditions.

\subsubsection{Generalization Across Different Garments}
To evaluate the generalization capability of our proposed scheme across different garments, we conducted alignment experiments on four garments. 
For this evaluation, we selected garments with visible patterns and used four feature points on each pattern. 
We performed ten trials for each garment and evaluated the results. 
The garments used in this experiment are shown in Fig.~\ref{fig:cloth}(b)-(e). 
T4, a long-sleeved T-shirt, is a different type of garment from the short-sleeved T-shirt mesh model used during training. 
Note that our Seam-to-Graph network is trained only on a simulated dataset, thus, all of the garments T0-T4 in Fig.~\ref{fig:cloth} are unseen garments by the network. 
The distribution of the feature-point RMSE is presented in Fig.~\ref{fig:compare}(b). 
Fig.~\ref{fig:cloth} shows the best and worst qualitative alignment results for each garment in this experiment in the second and third rows, respectively. 
For each case, we capture an image of the aligned garment pattern and superimpose it onto an image of the desired configuration.
The experimental results validate that our proposed method can generalize effectively to different garments, achieving alignment quality at or better than the level of humans.

\subsubsection{Robustness to Initial Grasping Offsets}
\begin{figure}[t]
    \includegraphics[width=\linewidth]{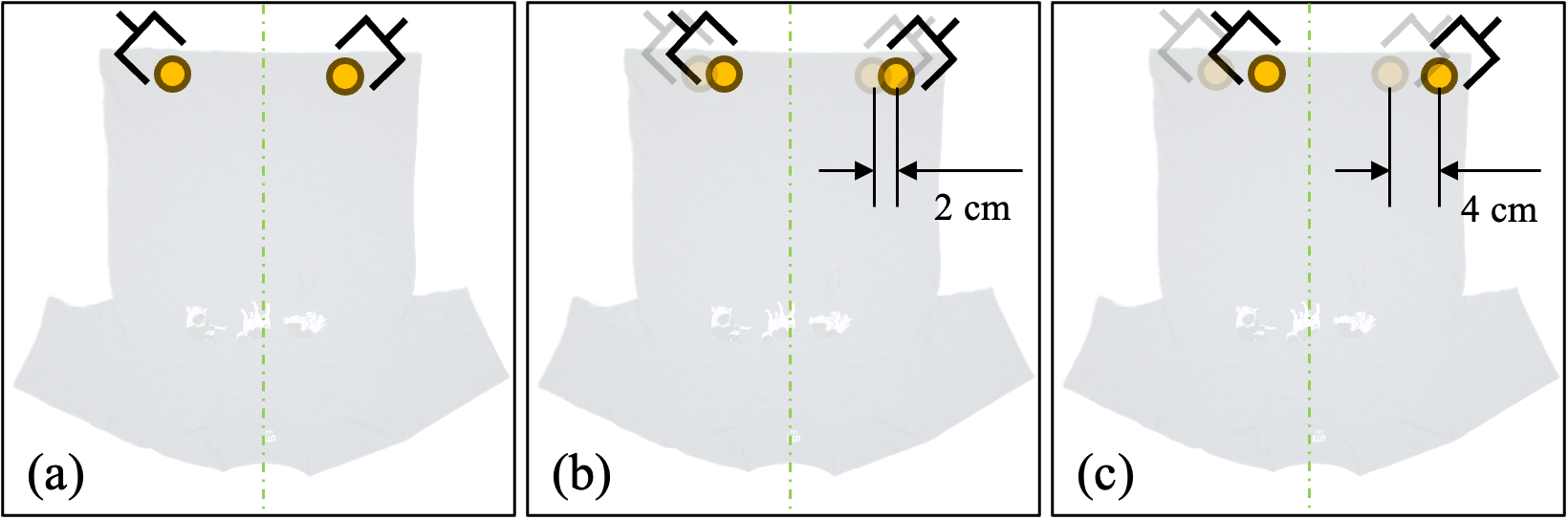}
    \caption{
        Illustration of initial grasping positions. 
        Yellow points indicate initial grasping points. 
        The green dashed line shows the garment's symmetry axis. 
        (a) shows the two grasping points in symmetric positions. 
        (b) and (c) show the grasping points shifted 2 and 4 cm, respectively, from the symmetric position.
    }
    \label{fig:graspbias}
\end{figure}

To further evaluate the system’s robustness, we conducted quantitative comparison experiments under different initial grasping conditions. 
As shown in Fig.~\ref{fig:graspbias}, in addition to symmetrical grasping (0 cm offset), we performed alignment experiments with offsets of 2 and 4 cm. 
For each initial state, we conducted ten trials using the garment shown in Fig.~\ref{fig:cloth}(b). 
We performed two-sided Welch's t-tests to compare the RMSE of the feature points between the different initial grasping offsets.
Table~\ref{tab:ttest} summarizes the results, and no statistically significant differences were observed between the groups.
These results demonstrate that our proposed control strategy can effectively correct initial state offsets, thereby reducing the requirements for upstream grasping tasks.

\begin{table}[t]
    \centering
    \caption{T-test results for different initial grasping offsets.}
    \label{tab:ttest}
    \small
    \setlength{\tabcolsep}{4.0pt}
    \begin{tabular}{lccccc}
        \hline
        \textbf{Offset} & \textbf{t-value} & \textbf{p-value} & \textbf{Significance} \\
        \hline
        0 cm vs. 2 cm & -1.719 & 0.1047 & n.s. \\
        0 cm vs. 4 cm & -0.988 & 0.3399 & n.s. \\
        2 cm vs. 4 cm & 0.415 & 0.6835 & n.s. \\
        \hline
    \end{tabular}
    {\begin{tablenotes}
    \item[*] n.s.: not significant ($\text{p-value} \geq 0.05$)~
    \end{tablenotes}}
\end{table}

\subsection{Discussion and Limitations}

The experimental results verify the robustness and feasibility of the proposed scheme.
While the system generalizes well to long-sleeved T-shirts, it remains a challenge to apply the seam-based perception and control method to garments with extreme deformability (e.g., sheer silk) or completely seamless structures. 
Additionally, implementing a more efficient code, such as using parallel computing for image and point cloud preprocessing, can increase the frequency of the entire method pipeline, achieving better real-time control performance.

\section{Conclusion}\label{sec:conclusion}

In this paper, we present a robust scheme for manipulating deformable garments, with a focus on the challenging tasks of automatically loading and aligning them.

To bridge the gap between unstructured visual observations and precise robotic control, we propose a Seam-to-Graph network that explicitly fuses heterogeneous multi-view point clouds and seam information into a fixed-topology structural skeleton graph.
Using structural priors, the perception module achieves robust and accurate estimation of the garment's state.

Based on the estimated structural skeleton graph, we introduce a deformation-aware visual servoing controller that aligns local garment regions iteratively using planar $\mathbb{SE}(2)$ registration and position-based visual servoing.

Extensive real-world experiments demonstrate that the proposed scheme achieves human-level alignment accuracy with improved repeatability, exhibits strong generalization across various garments, and is highly robust to initial grasping bias.



\bibliographystyle{IEEEtran}
\bibliography{ref.bib}
\vspace{-8mm}
\begin{IEEEbiography}
[{\includegraphics[width=1in,height=1.25in,clip,keepaspectratio]
{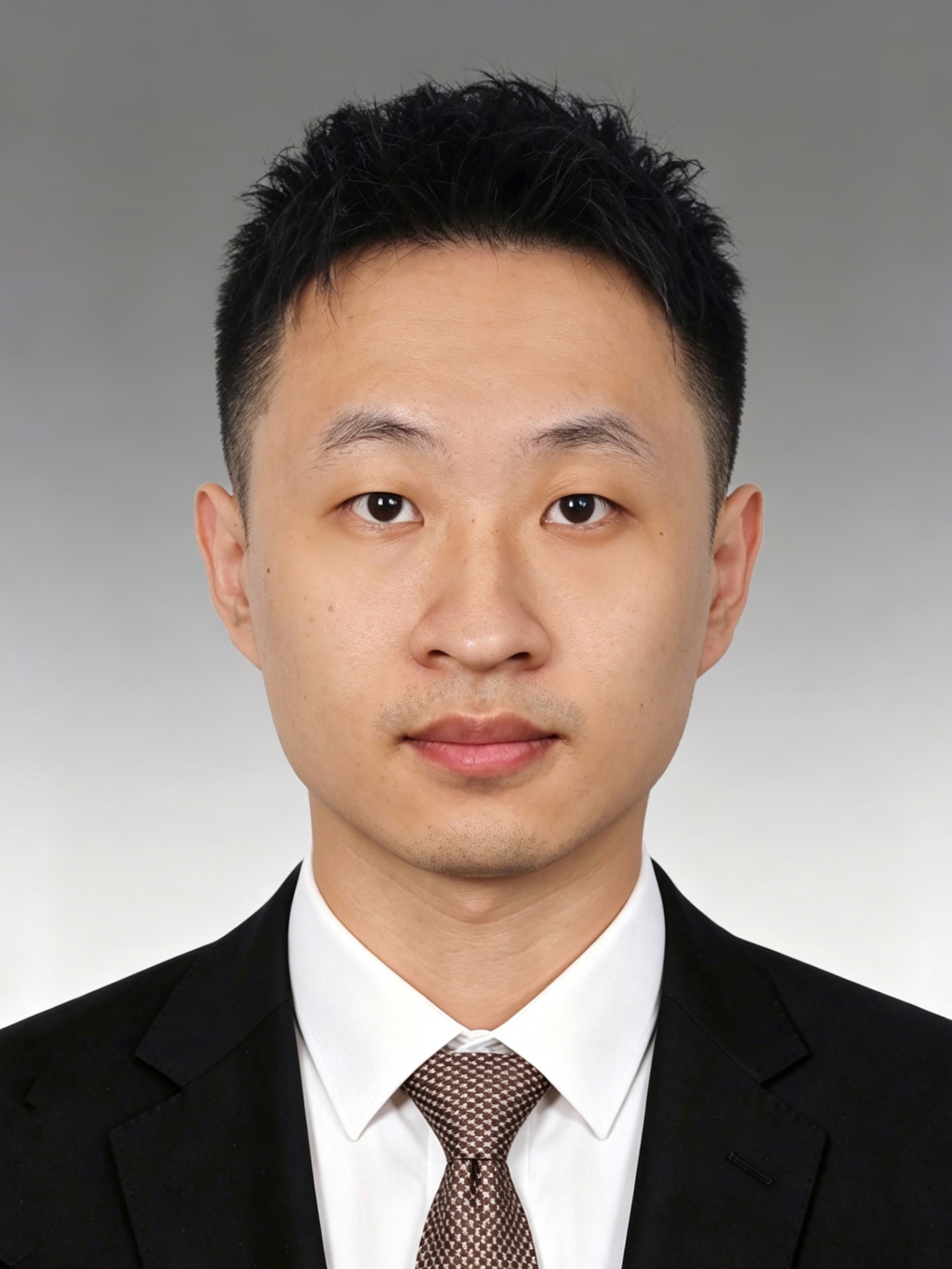}}]{Xuzhao Huang} (Graduate Student Member, IEEE) received his B.Eng. degree in mechanical design, manufacturing, and automation from Xiamen University, China, in 2018, and the M.Eng. degree in mechatronics engineering from the Harbin Institute of Technology, Shenzhen, in 2021. He is currently pursuing the Ph.D. degree in engineering with The University of Hong Kong. His research interests include robotic manipulation of deformable objects and visual perception.
\end{IEEEbiography}
\vspace{-8mm}

\begin{IEEEbiography}
[{\includegraphics[width=1in,height=1.25in,clip,keepaspectratio]
{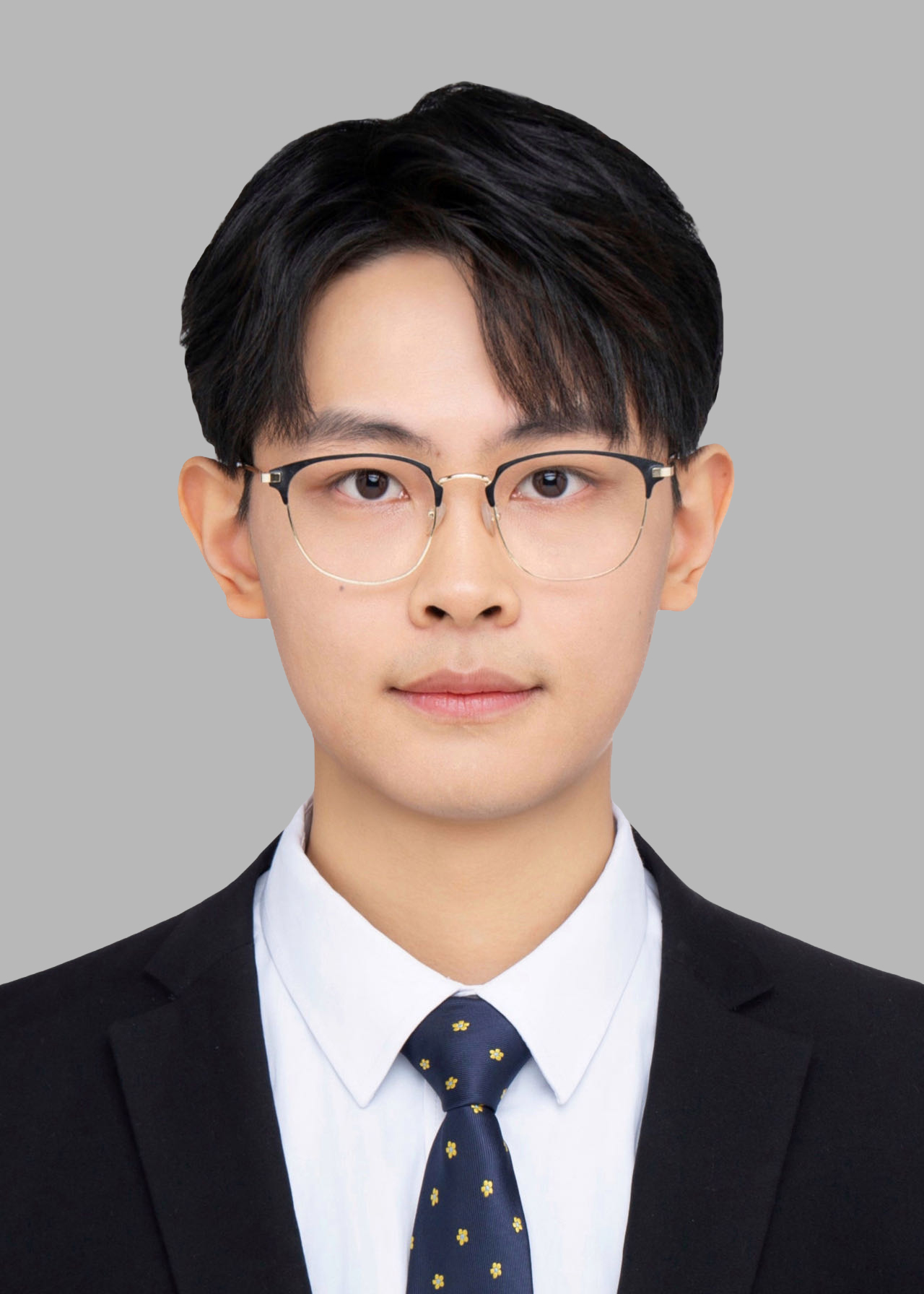}}]{Kai Tang} (Graduate Student Member, IEEE) received his B.Eng. degree in Process Equipment and Control Engineering from South China University of Technology in 2020, and M.Sc. degree with Distinction in Control Systems from Imperial College London in 2021. He is currently pursuing his Ph.D. degree in robotics at the JC STEM Lab of Robotics for Soft Materials, The University of Hong Kong. From 2022 to 2025, he worked with the Centre for Transformative Garment Production, Hong Kong SAR, which was in collaboration with Tohoku University, Japan. His research focuses on robot learning and control for fabric manipulation and fixture-free automated sewing.
\end{IEEEbiography}

\begin{IEEEbiography}
[{\includegraphics[width=1in,height=1.25in,clip,keepaspectratio]
{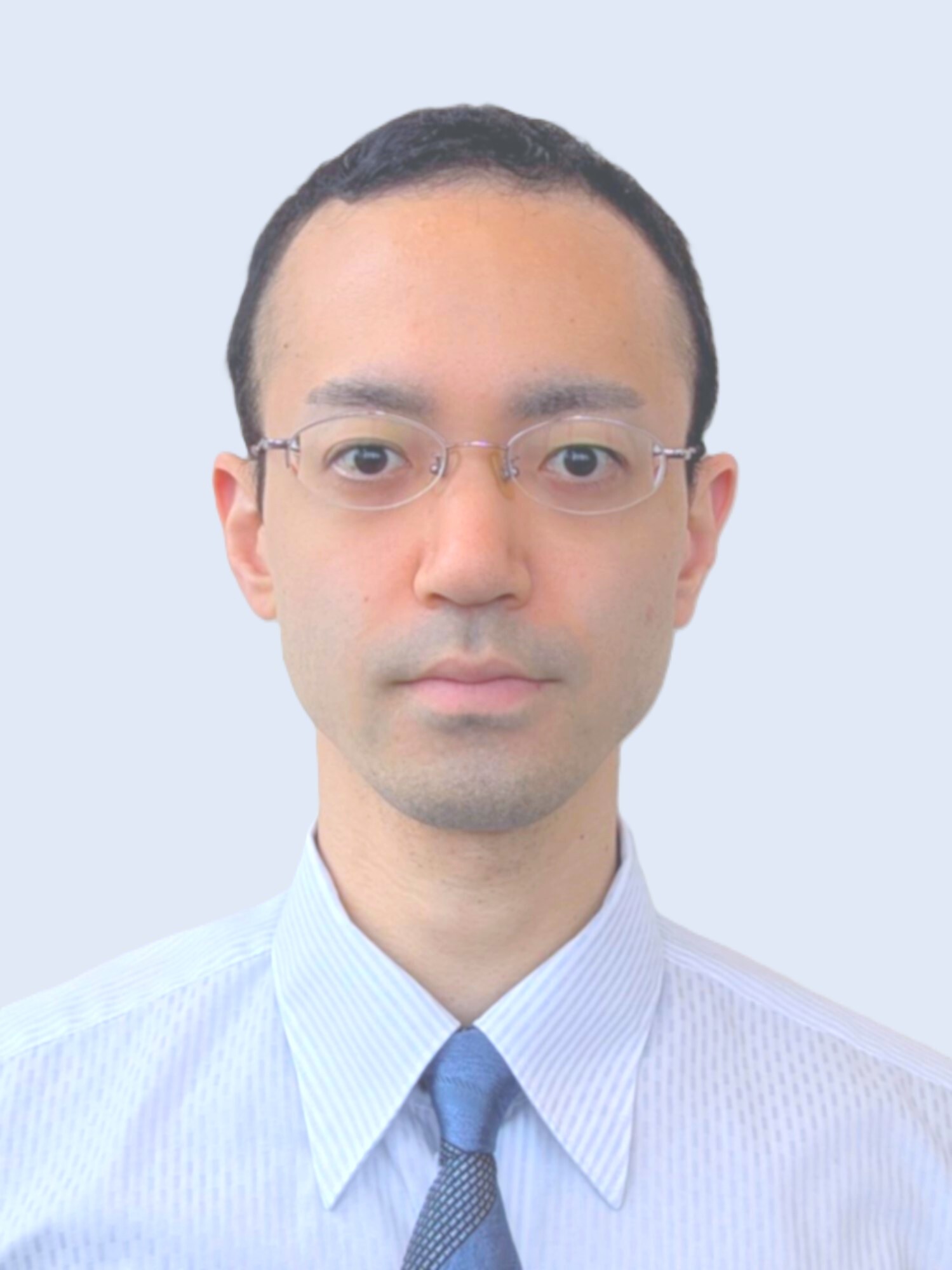}}]{Fuyuki Tokuda }
(Member, IEEE) received the B.S. degree in engineering from Nagoya Institute of Technology, Nagoya, Japan, in 2017, and the M.S. and Ph.D. degrees in engineering from Tohoku University, Sendai, Japan, in 2019 and 2022, respectively. From 2022 to 2023, he was a Post-Doctoral Fellow with the Centre for Transformative Garment Production (TransGP), an InnoHK Research Centre, jointly established by The University of Hong Kong and Tohoku University, Hong Kong SAR. From 2023 to 2025, he was a Research Officer with TransGP. He was a Visiting Research Associate with The University of Hong Kong from 2022 to 2025. Since 2025, he has been an Assistant Professor with the Unprecedented-Scale Data Analytics Center (UDAC), Tohoku University, and is also affiliated with the Graduate School of Information Sciences, Tohoku University. He was a recipient of the Research Fellowship from Tohoku University Graduate Program for Integration of Mechanical Systems in 2028 and the Research Fellowship from Japan Society for the Promotion of Science (JSPS) in 2021.
\end{IEEEbiography}

\begin{IEEEbiography}[{\includegraphics[width=1in,height=1.25in,clip,keepaspectratio]
{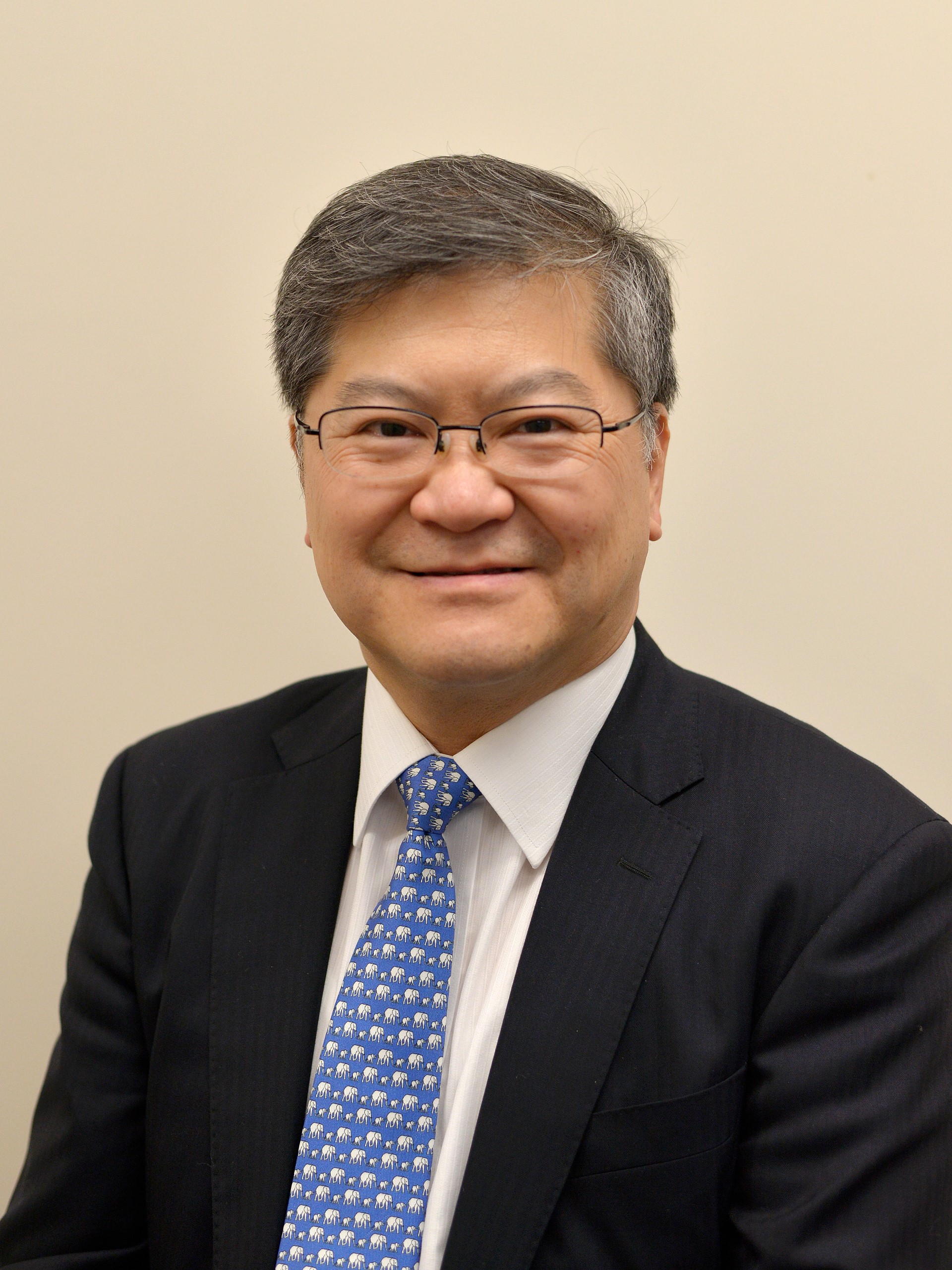}}]{Norman C. Tien} (Senior Member, IEEE) is the Taikoo Professor of Engineering and Chair Professor of Microsystems Technology at the University of Hong Kong (HKU). He is also currently the Head of Innovation Academy of Faculty of Engineering and the Managing Director of the Centre for Transformative Garment Production. He served as the Dean of Engineering from 2012 to 2018, and as the Vice-President and Pro-Vice-Chancellor (Institutional Advancement) from 2019 to 2021 at HKU.

Prior to joining HKU, Professor Tien was the Nord Professor of Engineering at Case Western Reserve University, where he was the Dean of Engineering from 2007 to 2011. He previously held faculty positions at University of California at Davis, University of California at Berkeley and Cornell University.

Professor Tien received his Ph.D. from the University of California at San Diego, MS from the University of Illinois, and BS from the University of California at Berkeley. His research interests are in the area of micro and nanotechnology, microelectromechanical (MEMS) systems, and robotics.

\end{IEEEbiography}
\begin{IEEEbiography}[{\includegraphics[width=1in,height=1.25in,clip,keepaspectratio]
{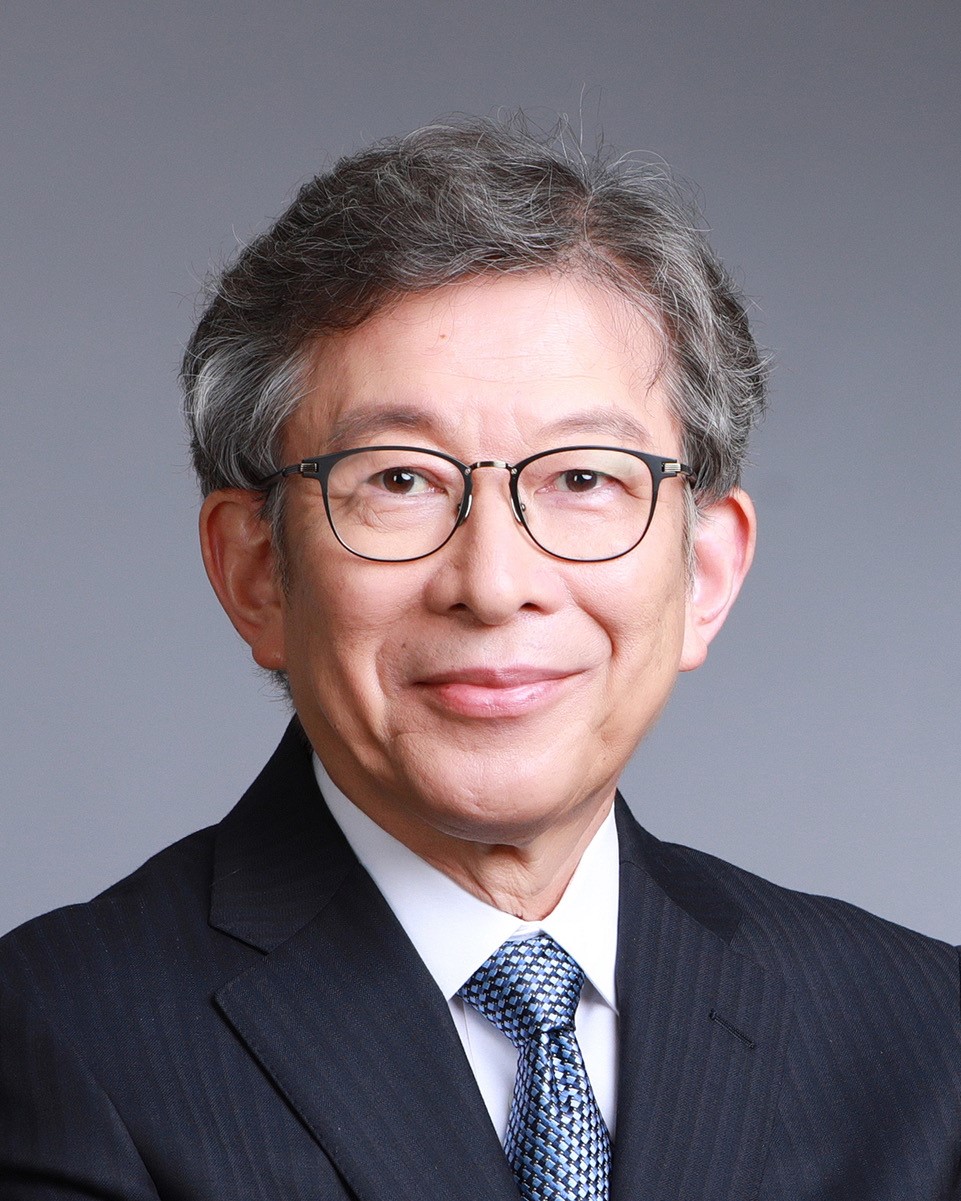}}]{Kazuhiro Kosuge} (Life Fellow, IEEE) received the B.S., M.S., and Ph.D. in control engineering from the Tokyo Institute of Technology, in 1978, 1980, and 1988 respectively. After having served as a R\&D Staff of the Production Engineering Department, Nippon Denso Company, Ltd., a Research Associate at Tokyo Institute of Technology, and an Associate Professor at Nagoya University, he joined Tohoku University as Professor in 1995 and served as Distinguished Professor from 2018 to March 2021. He is currently a Deputy Managing Director of the Centre for Transformative Garment Production, Hong Kong SAR, and the Director of the JC STEM Lab of Robotics for Soft Materials, Department of Electrical and Electronic Engineering, the University of Hong Kong, Hong Kong SAR.
    
He received the Medal of Honor, Medal with Purple Ribbon, from the Government of Japan in 2018 - a national honor in recognition of his prominent contributions to academic and industrial advancements. He also received IEEE RAS George Saridis Leadership Award in Robotics and Automation in 2021 for his exceptional vision of innovative research and outstanding leadership in the robotics and automation community through technical activity management. He is an IEEE Fellow, JSME Fellow, SICE Fellow, RSJ Fellow, JSAE Fellow and a member of the Engineering Academy of Japan. He was the President of the IEEE Robotics and Automation Society, from 2010 to 2011, the IEEE Division X Director, from 2015 to 2016, and the IEEE Vice President for Technical Activities for 2020.
\end{IEEEbiography}

\end{document}